\documentclass[final,5p,twocolumn]{elsarticle}
\usepackage[flushleft]{threeparttable}
\usepackage{graphicx}
\usepackage{makeidx} % allows for indexgeneration
\usepackage{amsmath}
\usepackage{array}
\usepackage{soul}

\usepackage{multirow}% http://ctan.org/pkg/multirow 
\renewcommand*{\thefootnote}{\fnsymbol{footnote}}
\usepackage{amsfonts}%
\usepackage{amssymb}%
\usepackage[ruled,vlined]{algorithm2e}
\usepackage{etoolbox}
\makeatletter
\patchcmd{\algocf@Vline}{\vrule}{\vrule\hspace{-0.25em}}{}{}
\makeatother
\usepackage{graphics}
\usepackage{subfig}
\usepackage{lineno}

\usepackage{setspace}
\usepackage{geometry}
\makeatletter
\g@addto@macro\endfrontmatter{\enlargethispage{-2\baselineskip}}
\makeatother

\usepackage[labelfont=bf]{caption}
\usepackage[font={small}]{caption}
\usepackage{array}
\usepackage{makecell}

\usepackage{transparent}
\usepackage{color}
\usepackage{xcolor}
\usepackage[utf8]{inputenc}
\usepackage[english]{babel}
\usepackage[hidelinks]{hyperref}

\makeatletter
\newcommand\footnoteref[1]{\protected@xdef\@thefnmark{\ref{#1}}\@footnotemark}
\makeatother
\usepackage{tabularx}
\usepackage{enumitem}
\usepackage{adjustbox}

\usepackage[figuresright]{rotating}

\setcitestyle{square}

\journal{NeuroImage}

\begin{document}

\begin{frontmatter}

\title{Automated Olfactory Bulb Segmentation on High Resolutional T2-Weighted MRI}
\author[label1,label2]{Santiago Estrada}
\author[label2]{Ran Lu}
\author[label1]{Kersten Diers}
\author[label2]{Weiyi Zeng}
\author[label3]{Philipp Ehses}
\author[label3,label4]{Tony St\"ocker}
\author[label2,label5]{Monique M.B Breteler}
\author[label1,label6,label7]{Martin Reuter\corref{cor1}}

\address[label1]{Image Analysis,German Center for Neurodegenerative Diseases (DZNE), Bonn, Germany}
\address[label2]{Population Health Sciences,German Center for Neurodegenerative Diseases (DZNE), Bonn, Germany}
\address[label3]{MR Physics,German Center for Neurodegenerative Diseases (DZNE), Bonn, Germany}
\address[label4]{Department of Physics and Astronomy, University of Bonn, Germany}
\address[label5]{Institute for Medical Biometry, Informatics and Epidemiology (IMBIE), Faculty of Medicine, University of Bonn, Bonn, Germany}
\address[label6]{A.A. Martinos Center for Biomedical Imaging, Massachusetts General Hospital, Boston MA, USA }
\address[label7]{Department of Radiology, Harvard Medical School, Boston MA, USA}

\cortext[cor1]{Correspondence to: Martin Reuter (\texttt{martin.reuter [at] dzne.de}).}

\begin{abstract}
The neuroimage analysis community has neglected the automated segmentation of the olfactory bulb (OB) despite its crucial role in olfactory function. The lack of an automatic processing method for the OB can be explained by its challenging properties (small size, location, and poor visibility on traditional MRI scans). Nonetheless, recent advances in MRI acquisition techniques and resolution have allowed raters to generate more reliable manual annotations. Furthermore, the high accuracy of deep learning methods for solving semantic segmentation problems provides us with an option to reliably assess even small structures.  In this work, we introduce a novel, fast, and fully automated deep learning pipeline to accurately segment OB tissue on sub-millimeter T2-weighted~(T2w) whole-brain MR images. To this end, we designed a three-stage pipeline: (1) Localization of a region containing both OBs using \textit{FastSurferCNN}, (2) Segmentation of OB tissue within the localized region through four independent \textit{AttFastSurferCNN} - a novel deep learning architecture with a self-attention mechanism to improve modeling of contextual information, and (3) Ensemble of the predicted label maps. For this work, both OBs were manually annotated in a total of 620 T2w images for training (n=357) and testing.
The OB pipeline exhibits high performance in terms of boundary delineation, OB localization, and volume estimation across a wide range of ages in 203 participants of the Rhineland Study (Dice Score (Dice): 0.852, Volume Similarity (VS): 0.910, and Average Hausdorff Distance (AVD): 0.215~$mm$). 
Moreover, it also generalizes to scans of an independent dataset never encountered during training, the Human Connectome Project~(HCP), with different acquisition parameters and demographics, evaluated in 30 cases at the native  0.7~$mm$ HCP resolution (Dice: 0.738, VS: 0.790, and AVD: 0.340~$mm$), and the default 0.8~$mm$ pipeline resolution (Dice: 0.782, VS: 0.858, and AVD: 0.268~$mm$). We extensively validated our pipeline not only with respect to segmentation accuracy  but also to known OB volume effects, where it can sensitively replicate age effects ($\beta = - 0.232$, $p<0.01$). Furthermore, our method can analyze a 3D volume in less than a minute (GPU) in an end-to-end fashion, providing a validated, efficient, and scalable solution for automatically assessing OB volumes.
\end{abstract}


\end{frontmatter}

%\linenumbers

\section{Introduction}
\subsection{Motivation}
Over the past decades, there has been an increasing awareness to odor function not only as a quality of life indicator~\cite{lifequality} but also as a potential biomarker in population studies. Olfactory dysfunction is among the earliest signs of many neurodegenerative disorders, including Alzheimer's and Parkinson's disease~\cite{attems2014olfactory,doty2017olfactory,roberts2016association}. Therefore, it is of major interest to gain insights into the anatomical basis of the olfactory pathway \textit{in vivo}.

New developments in magnetic resonance imaging (MRI) (e.g.\ field strength, accelerated acquisition schemes, etc.) have allowed the acquisition of high-resolutional (High-Res) MR images, providing an option for reliable assessment of odor-related brain structures, including olfactory bulb (OB). The OB is considered the most important relay station in the odor pathway, integrating peripheral and central olfactory information. Moreover, OB volume has been associated with olfactory dysfunction in clinical settings~\cite{ob_strufun, hummel2011correlation}. However, compared to its central counterparts, i.e.\ prefrontal cortex, hippocampus, and insular cortex~\cite{Dintica2019, Vassilaki2017}, OB remains relatively poorly studied, especially in the general population. One reason for that could be the lack of a fully automated segmentation tool for this structure.

Currently, the gold standard for measuring OB volumes is the manual segmentation of T2 weighted (T2w) images --a very 
expensive and time-consuming process that greatly relies on the raters' expertise. Thus, especially for large population-based studies, automatic segmentation methods are required. However, achieving good accuracy on this small structure is challenging due to its inherent properties: (i) low contrast on T1w scans, (ii) low boundary contrast on T2w images (partial volume effects), (iii) highly sensitivity to noise due to its proximity to the nostrils (e.g.\ breathing artefacts), (iv) not visible in all subjects~\cite{weiss2020human}, and (v) highly dependent of age~\cite{hummel2011correlation,buschhuter2008correlation,hummel2015volume}. So far, those limitations have impeded the wide implementation of any automatic or semi-automatic techniques. Therefore, the introduction of an accurate automated method for segmenting OB is of significant clinical and research interest.

\subsection{Olfactory Bulb Segmentation}
Despite the fact, that many studies have analyzed the OB, there is a lack of accurate automatic processing methods for this structure which has been overlooked by many of the standard neuroimage processing frameworks, such as \textit{FreeSurfer}~\cite{freesurfer1}, \textit{BrainSuite}~\cite{shattuck2002brainsuite}, \textit{SPM}~\cite{friston2003statistical}, \textit{ANTs}~\cite{avants2009advanced}, or \textit{FSL}~\cite{JENKINSON2012782}.
To date, manual delineation is still the predominant approach for accurate quantification of OB volumes. Most groups approximate OB volumes from 1.5T T2w MR scans with a relative low resolution (of 1.5~$mm$ to 2~$mm$ isotropic) ~\cite{hummel2011correlation,buschhuter2008correlation,hummel2015volume,cortex}. Recent studies~\cite{weiss2020human, obsemi} on 3T high-resolutional T2w MRI have focused on developing semi-automatic techniques to reduce manual annotations workload but cannot automatically segment the OB. Concurrently to our work, \textit{Noothout et al.}~\cite{noothout2021automatic} proposed an automatic pipeline using fully convolutional neural networks (F-CNNs) to segment the OB on coronal T2w images with an in-plane resolution of 0.47$~mm~\times$~0.47~$mm$ and 1~$mm$ thickness. While this method, which is not publicly available at this time, shows promising results in a small dataset (n=21), it is reported to be sensitive to motion artefacts and unseen scenarios (i.e.\ cases with no apparent OB). 

Recently, supervised learning using F-CNNs~\cite{segnet,FCNNS} has become the preferred standard in the medical computer vision community for solving semantic segmentation problems when sufficient training data is available~\cite{noothout2021automatic,quicknat,kamnitsas2017efficient,billot2020automated,ronneberger2015u,milletari2016v,dong2017automatic,roy2018recalibrating,estrada2020fatsegnet,henschel2020fastsurfer}. F-CNNs often outperform other traditional methods, as they can learn intrinsic features and integrate global context to resolve local ambiguities in an end-to-end fashion. The most frequently employed network layout for semantic segmentation is the encoder-decoder architecture, i.e.\ the \textit{UNet}~\cite{ronneberger2015u}. The accuracy of this architecture, however, decreases when segmenting smaller structures~\cite{billot2020automated,roy2018recalibrating,competitionvsconcatenation}. This can be due to the more complex shapes (i.e.\ thinner, irregular boundaries) and visual appearance characteristics in medical images (i.e.\ less visible and partly occluded). Nonetheless, some of the fault can be attributed to the encoder-decoder layout as it can lead to a redundant use of information and insufficient encoding of the global contextual information~\cite{fu2019dual,sinha2020multi}. An accurate understanding of the spatial context is of tremendous importance when segmenting smaller structures as local representation differences between pixels/voxels of a same structure introduce inter-class inconsistencies and affect the recognition accuracy~\cite{fu2019dual}. To solve this issue, attention modules have been introduced to improve the understanding of long-range dependencies, not only for semantic segmentation~\cite{roy2018recalibrating,fu2019dual,sinha2020multi} but also for other computer vision tasks~\cite{zhang2019self,lin2016efficient,lin2017focal,vaswani2017attention}. 

In this work, we modify our \textit{FastSurferCNN}~\cite{henschel2020fastsurfer} for whole-brain segmentation to focus on the OB. To improve \textit{FastSurferCNN's} performance for small structures, we suitably included the self-attention mechanism proposed in \cite{zhang2019self} into \textit{FastSurferCNN}; the new deep-learning architecture is termed \textit{AttFastSurferCNN}. \textit{AttFastSurferCNN} promotes attention to spatial information by improving the modeling of local and global-range dependencies. Overall, to segment the OB on high-resolutional T2w whole-brain MRI in a fully automatic fashion, we introduce a deep learning pipeline consisting of three stages:

\begin{enumerate}
    \item {\bf Localization} of a region of interest (ROI) containing the OBs of both hemispheres using a semantic segmentation approach by implementing \textit{FastSurferCNN}; we use the centroid of the predicted region as a center point for cropping a localized volume.   
    \item {\bf Segmentation} of OB tissue within the localized volume through four \textit{AttFastSurferCNN} with different training condition (four data-splits and data initialization). 
     \item {\bf Ensemble} stage where the previously generated label maps are averaged and view-aggregated to form a consensual final segmentation.
\end{enumerate}

 \begin{figure*}[!hbt]
    \centering
    \includegraphics[width=0.8\textwidth]{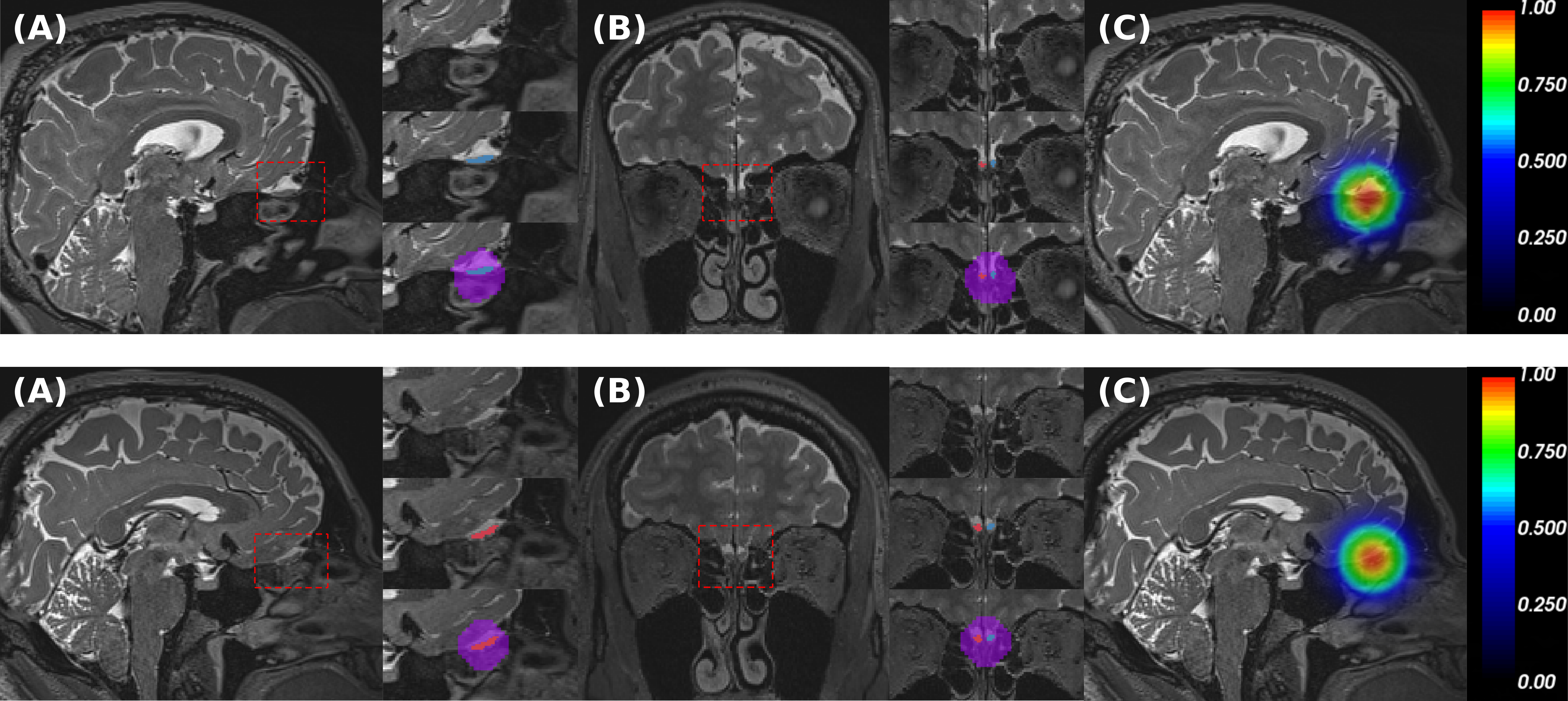}
    \caption{T2-weighted images and ground truth from two subjects. The red square represents the zoom-in region. A) Sagittal view and labels (blue: left OB, red: right OB, purple: ROI label). B) Coronal view and labels. C) ROI  distance map around the centroid of the OB labels.}
    \label{fig:manual_labels}
\end{figure*}

 \renewcommand{\thefootnote}{\fnsymbol{footnote}}
The presented networks were trained with manual annotations of 357 T2w scans from the Rhineland Study, an ongoing large population-based cohort study~\cite{breteler2014mri,stocker2016big}. We extensively validated the quality of the individual stages of the pipeline through assessment of segmentation accuracy in an independent unseen heterogeneous in-house dataset ($n=203$). We showed that our previously introduced \textit{FastSurferCNN} can precisely localize the region containing both OBs and that the proposed \textit{AttFastSurferCNN} can accurately segment the OBs, outperforming other establish F-CNNs and accomplishing equivalent results as manual raters. After asserting segmentation accuracy, we validated the soundness of the proposed pipeline in the Rhineland Study with respect to: i) replication of known OB volume effects (e.g.\ age), ii) stability of volume estimates among variations of the study's T2w sequences, and iii) robustness to scans without an apparent OB. We further assessed generalizability to an unseen externally labeled dataset of 30 subjects from a cohort with different characteristics and acquisition parameters. To the best of our knowledge, our pipeline is the first framework capable of automatically segmenting the OB in a large cohort dataset with high accuracy and reliability. Furthermore, we demonstrated that our method can generalize to different T2w scans with 0.8~$mm$ isotropic resolution. The proposed method is available as an open-source project at: https://github.com/Deep-MI/olf-bulb-segmentation.

\section{Methodology}

\subsection{Manual Reference Standard\label{sec:labels}}
\noindent
Our manual reference standard is based on the annotation of high-resolutional (0.8~$mm$ isotropic) T2w MRI from the Rhineland Study. The Rhineland Study (www.rheinland-studie.de/)  is an ongoing study that enrolls participants aged 30 years and above at baseline from Bonn, Germany. The study is carried out in accordance with the recommendations of the International Council for Harmonisation (ICH) Good Clinical Practice (GCP) standards (ICH‐GCP). Written informed consent was obtained from all participants in accordance with the Declaration of Helsinki.  

Manual annotations of the left and right OB were performed by an experienced rater in (unprocessed) T2w images using \textit{Freeview} (a visualization tool of \textit{FreeSurfer}~\cite{freesurfer1,freesurfer2}). OB is defined as a mostly almond- or spindle-shaped structure symmetrically located at the base of the forebrain~\cite{rombaux2009measure} as seen in Figure~\ref{fig:manual_labels}, which can be demarcated based on surrounding cerebrospinal fluid and the underlying cribriform plate. The abrupt changes in diameter at the beginning of the olfactory tract in the axial and sagittal views were used as a posterior ending landmark~\cite{wang2011association,yousem1998olfactory}. In addition, to avoid bias, labeling was blind to participant metadata, e.g.\ outcomes of the olfactory function and demographics.

\begin{figure*}[!hbt]
    \centering
    \includegraphics[width=0.9\textwidth]{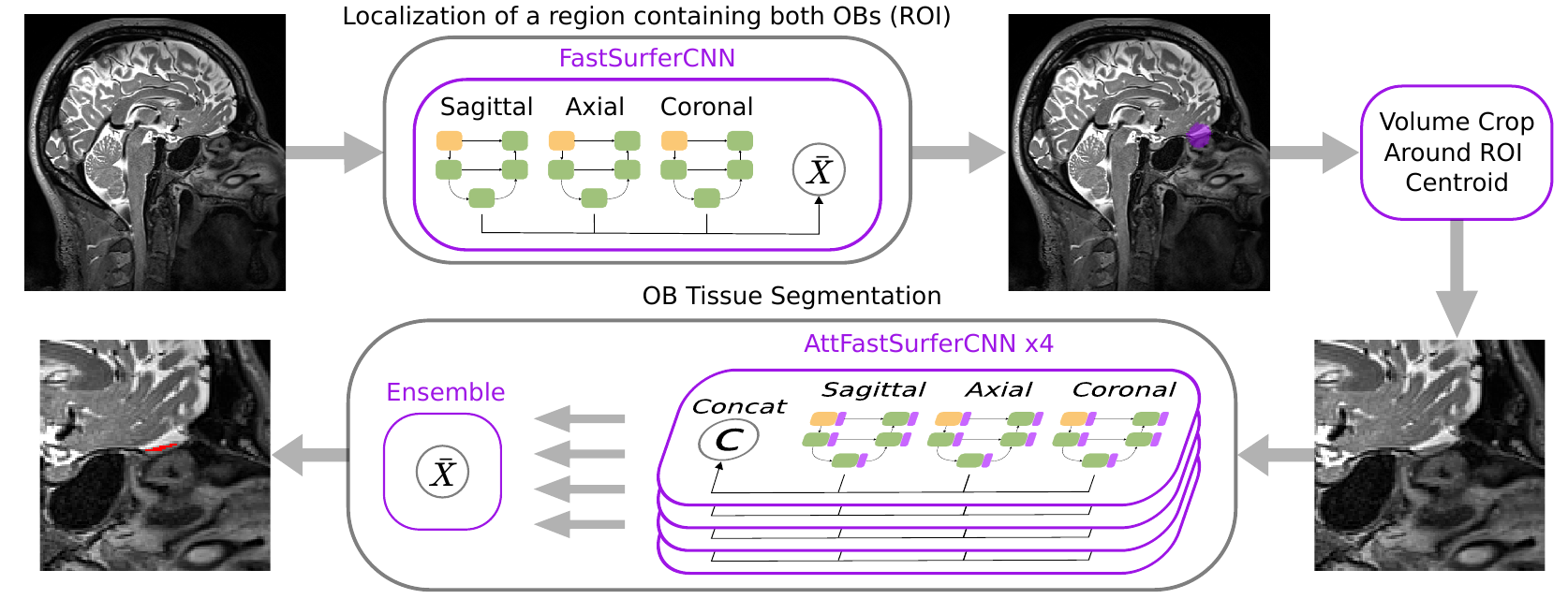}
    \caption{Proposed pipeline for OB segmentation. The pipeline is divided into three stages: First, localization of a region of interest containing the left and right OB. Then, OB tissue segmentation within the localized volume, and finally, an ensemble of predicted label maps.}
    \label{fig:obpipeline}
\end{figure*}

For the localization task, we solve a semantic segmentation problem with the goal to segment the forebrain region containing the OBs from both hemispheres (referred to as “region of interest (ROI)”). The ROI label generation is achieved by the following steps: (1) Localization of the mid-point between left and right OB by calculating the centroid ($C$) of the manual labels. (2) Generation of a distance map by applying a Gaussian distribution around  $C$ on a down-sampled 1.6~$mm$ isotropic image, the distance map is defined as : $f(x,y,z) = \mathcal{N}(\mu = C,\,\delta = 10)$ where x ,y and z are voxel coordinates in the down-sampled image. (3) A binary cutoff at $f(x,y,z) / max(f(x,y,z)) >= 0.8$ separates ROI and background. The Resulting distance maps and labels are illustrated in Figure~\ref{fig:manual_labels}.

\subsection{Olfactory Bulb Pipeline}

Our proposed deep learning method is aimed at segmenting the OB on high-resolutional T2w whole-brain MRI. This task presents the challenge of a high-class imbalance between foreground and background ($\approx 1:10^{6}$). A reduction in the spatial size of the input can partially alleviate the problem by cropping the background and by focusing the background information on relevant regions in close proximity to the OBs. This, furthermore, reduces computational and memory requirements during training and inference. Following this direction, we designed a fully automated pipeline for OB tissue segmentation as depicted in Figure~\ref{fig:obpipeline}.

The proposed pipeline consists of three stages: (1) In order to remove most of the unnecessary background we first train \textit{FastSurferCNN}~\cite{henschel2020fastsurfer} with a down-sampled 1.6~$mm$ isotropic image to provide a quick segmentation of the forebrain region containing both OBs (localization network). This segmentation is only used to compute a centroid coordinate of the region of interest. A final localized volume (at 0.8~$mm$ isotropic, 96~$\times$~96~$\times$~96~voxels), centered at this coordinate, is cropped or resampled from the input image. By default the pipeline resamples deviating resolutions to 0.8~$mm$ isotropic, unless the user specifies to use the native resolution instead. (2) Afterwards, the OB tissue is segmented within this cropped volume by four \textit{AttFastSurferCNNs} with different training conditions (four data-splits and data initialization). (3) Finally, the ensemble segmentation is composed by averaging the predicted label maps; the implemented ensemble approach ensures that only voxels with high agreement among models are selected and also reduces variance due to network initialization. Furthermore, since right and left OB were combined as one structure during segmentation, they are split retrospectively in an independent post-processing step. 

\subsubsection{ Region of Interest (ROI) Localization Network - FastSurferCNN}\label{localization}

To localize the ROI as a semantic segmentation task, we employ \textit{FastSurferCNN}~\cite{henschel2020fastsurfer}  as it outperformed other commonly used encoder-decoder architectures, i.e.\ \textit{SDNet}~\cite{sdnet} and \textit{QuickNat}~\cite{quicknat}, on the difficult task of whole-brain segmentation. \textit{FastSurferCNN} consists of three 2D F-CNNs operating on different anatomical views (coronal, axial, and sagittal) and a final view-aggregation stage. In brief, all F-CNNs follow the same layout of four competitive-dense blocks (CDB) for the encoder and decoder path separated by a bottleneck block. The use of CDB reduces the number of learnable parameters by replacing the typical concatenation units inside dense-connections with maxout activations~\cite{denseconnections,maxout}. The maxout activation induces competition between feature maps by computing the maximum at each spatial location, thus improving the feature selectivity~\cite{competition} and boosting the learning of fine-grained structures~\cite{estrada2020fatsegnet,competitionvsconcatenation}. Furthermore, \textit{FastSurferCNN}  utilizes a multi-slice input approach by stacking preceding slices, current, and succeeding slices for segmenting only the middle slice, which in turn increases the spatial information aggregation in a 2D network by improving the local neighborhood awareness~\cite{henschel2020fastsurfer}. 

\begin{figure}[!hbt]
    \centering
    \includegraphics[width=0.45\textwidth]{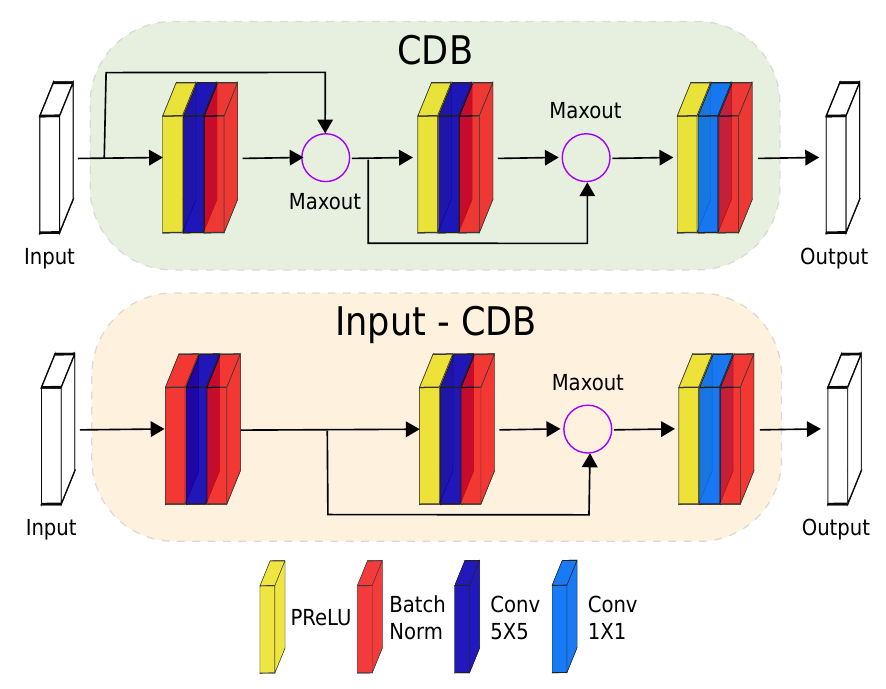}
    \caption{Competitive Dense Blocks (CDB) configuration. Each block is composed of three sequences of parametric rectified linear unit (PReLU), convolution (Conv) and batch normalization (BN) (bottom) with exception of the very first encoder block (top). In the first block, the PReLU is replaced with a BN to normalize the raw inputs.}
    \label{fig:CDB}
\end{figure}

In this work, we slightly modified \textit{FastSurferCNN} by adjusting the view-aggregation step to a normal unweighted average. Since the ROI label is not lateralized, there is no need to increase attention to any particular anatomical view. Furthermore, the prior downsampling of the input scan (to isotropic 1.6~$mm$) allows a reduction of the multi-slice input image from 7 to 3 consecutive slices while retaining approximately the same field of view. In terms of the CDB blocks, the three configuration sequences of a parametric rectified linear unit (PReLU), convolution (Conv)(64 filters), and batch normalization (BN) are maintained (Figure~\ref{fig:CDB} top) as well as the exception for the very first encoder block. In the first block, the first PReLU is replaced with a BN to normalize the raw inputs  (Figure~\ref{fig:CDB} bottom). 

 \begin{figure*}[!hbt]
    \centering
    \includegraphics[width=0.9\textwidth]{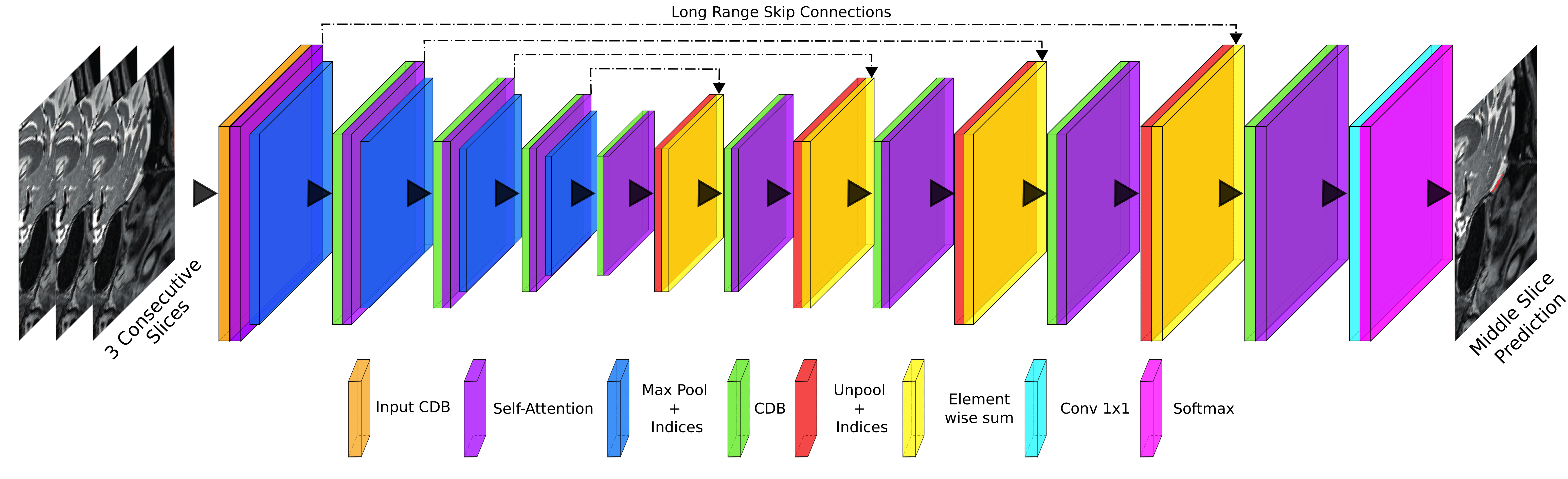}
    \caption{\textit{AttFastSurferCNN} network architecture. The network consists of four competitive dense blocks (CDB) in the encoder and decoder part, separated by a bottleneck layer. After each CDB a self-attention module is added. The CDB  configuration is illustrated in Figure~\ref{fig:CDB}.}
    \label{fig:attFastSurferCNN}
\end{figure*}

\subsubsection{OB Segmentation Network - AttFastSurferCNN}
To accurately segment the OB, we introduce \textit{AttFastSurferCNN} a new deep learning architecture that boosts the attention to spatial information. We implemented \textit{AttFastSurferCNN} by suitably including the self-attention mechanism proposed by~\cite{zhang2019self} into \textit{FastSurferCNN}~\cite{henschel2020fastsurfer}. The self-attention module was included  after each competitive-dense block(CDB), as shown in Figure~\ref{fig:attFastSurferCNN}, thus improving the modeling of contextual information. Furthermore, in order to take full advantage of the multi-scale attention maps~\cite{fu2019dual,sinha2020multi} and to prevent information loss from the unpooling layers~\cite{competitionvsconcatenation}, we replaced the maxout activation units between the finer feature maps from long-range skip connections and the coarser feature maps from the unpooling path with an element-wise sum. 

\begin{figure}[!hbt]
    \centering
    \includegraphics[width=0.48\textwidth]{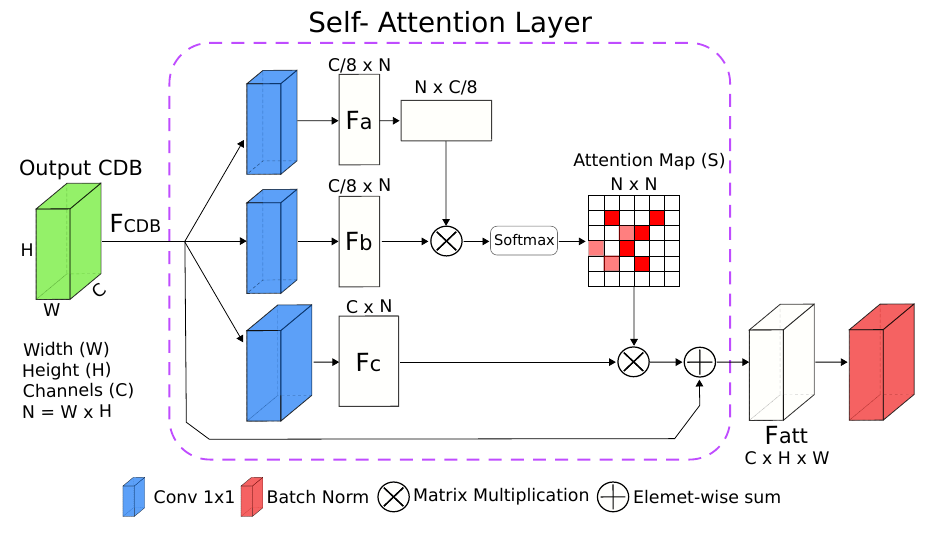}
    \caption{Implemented Self-Attention module within \textit{AttFastSurferCNN}}
    \label{fig:selfatt}
\end{figure}

The implemented self-attention layer is illustrated in Figure~\ref{fig:selfatt}. Let us denote the CDB output feature map as $F_{CDB}\:\epsilon\:\mathbb{R}^{C \times H \times W}$, where $C,H,W$ represent the channel, height, and width dimensions respectively. First, the $F_{CDB}$ is fit into two $1 \times 1$ convolutional layers to reshape the channels to a size of $C/8$ and create two new features maps ($F_{a}$ and $F_{b}$). Reducing the number of channels drastically diminishes memory requirements without a significant performance loss~\cite{zhang2019self}. Subsequently, the feature maps are flattened to a shape of $(C/8) \times (N)$, where $N=H \times W$ are the number of pixels. Afterwards, an attention map ($\textbf{S}$) is created by applying a softmax layer into the output of a matrix multiplication between $F_{a}^\top$ and $F_{b}$. Thus $\textbf{S}\:\epsilon\:\mathbb{R}^{N \times N}$ is defined as:

\begin{equation}
\begin{aligned}
s_{j,i} = exp\left ( F_{ai}^\top \cdot F_{bj} \right ) / \sum_{i=1}^{N} exp\left ( F_{ai}^\top \cdot F_{bj} \right )
\label{eq:1}
\end{aligned}
\end{equation}
 where $s_{j,i}$ indicates the extend to which the $i_{th}$ position impacts the $j_{th}$ position. Before applying $\textbf{S}$, the $F_{CDB}$ features are fed into a $1 \times 1$ convolutional layer and a new feature map $F_{c}\:\epsilon\:\mathbb{R}^{C \times H \times W}$ is generated and reshaped to $\mathbb{R}^{C \times N}$. Afterwards, a matrix multiplication is performed between the transpose of $\textbf{S}$ and $F_{c}$  and the results reshaped to the original size $\mathbb{R}^{C \times H \times W}$. Finally, the self-attention output ($F_{att}$) is formulated as follows:
 
 \begin{equation}
\begin{aligned}
F_{att} = \alpha(\textbf{S}^\top\cdot F_{c}) + F_{CDB}
\label{eq:2}
\end{aligned}
\end{equation} 

where $\alpha$ is a learnable scalar parameter initialized with 0. The introduction of $\alpha$ allows the network to first focus on the local information which is an easier task and gradually increases the importance of non-local dependencies which is a harder task~\cite{zhang2019self}.
We additionally normalize $F_{att}$ thus guaranteeing a normalized input to the other CDB blocks. A normalized input improves convergence~\cite{liao2016importance} and increases the exploratory span of the created sub-networks when using a maxout activation~\cite{competition}. In summary, the implemented spatial attention module improves the modelling of local and global-range dependencies, which in turn increases semantic consistency.

In brief, \textit{AttFastSurferCNN} is a multi-network approach of three 2D F-CNNs operating on different anatomical views (coronal, sagittal and, axial). All three F-CNNs contain the self-attention layers following the aforementioned layout (Figure~\ref{fig:attFastSurferCNN}). Within \textit{AttFastSurferCNN} the CDB blocks maintain the configuration from Section~\ref{localization} except for the $5\times5$ convolutions that are modified to a smaller kernel size of $3\times3$. Furthermore, the multi-slice input approach from \textit{FastSurferCNN}~\cite{henschel2020fastsurfer} is maintained and a stack of three consecutive slices are passed as input. In the following section, the ensemble of different segmentation predictions will be explained in detail.    

\subsubsection{OB Segmentation Ensemble}\label{sec:ensemble}
One widely used method to assess the optimal model of CNNs trained with different data-splits is cross-validation. Cross-validation jointly evaluates performance on different data-splits and the model with the maximal test-set performance is selected as the winner. This approach, however, can limit generalizability as the data-splits used for training the best performer can be biased towards the selected test-set.
Recently, the combination of different CNN model outputs has been shown to improve the prediction performance and reduce the CNN's intrinsic variance~\cite{ju2018relative}. As a consequence, we propose to ensemble the prediction of four \textit{AttFastSurferCNNs} trained with different data-splits, ensuring that only OB voxels with a high inter-model agreement are segmented, and thus reducing the bias to any particular data division. To ensure that all networks have a comparable OB knowledge: i) training was done under the same learning conditions (i.e.\ number of epochs, batch size, loss function, a learning rate scheduler, etc.), ii) training data was divided into four data-splits balanced for age and sex, and iii) the data-splits were treated in a leave-one-out fashion. Finally, the ensemble is constructed by an unweighted average as the output of models with comparable performance is merged~\cite{ju2018relative,he2016deep,szegedy2015going,kamnitsas2017ensembles}. Intuitively, the proposed ensemble approach can be seen as four different raters with similar experience taught by the same instructor and the consensus among the raters gives the final decision.  It is important to note, that in our specific approach the final ensemble prediction is created by averaging twelve different models as each \textit{AttFastSurferCNN} contains three 2D F-CNNs for the three different anatomical views (axial, coronal and, sagittal). Therefore, our ensemble approach also includes the advantages of view-aggregation where a voxel prediction is regularized by considering spatial information from multi-views~\cite{quicknat,estrada2020fatsegnet,henschel2020fastsurfer}. We furthermore analyzed the impact of the ensemble approach by comparing directly with standalone data-splits.

\subsubsection{Model Learning}\label{sec:model_learning}
All F-CNN models for localization and segmentation were implemented in PyTorch~\cite{paszke2017automatic} using a docker container~\cite{merkel2014docker}. Independent models for axial, coronal, and sagittal views were trained for 40 epochs with a batch size of 16 using two NVIDIA Tesla V100 GPU with 32 GB RAM, and a Adam optimizer~\cite{adamOptimizer} with a step decay scheduler that decreases the learning rate~(lr) by 95\% every 5 epochs (initial lr = 0.01, constant weight decay = $10^{-04}$~\cite{loshchilov2018decoupled}, betas=(0.9, 0.999), eps=$10^{-08}$). The networks were trained by optimizing a composed loss function of  focal loss~\cite{lin2017focal} and dice loss~\cite{milletari2016v}. The focal loss addresses the class imbalance by modifying the standard cross-entropy loss such that lower importance is given to the well-classified pixels. On the other hand, the dice loss is a more robust loss to handle data imbalance~\cite{sudre2017generalised} as it is based on the Dice score, an overlay similarity index
that reflects both size and localization agreement. Therefore, our proposed composed loss function is formulated as:
\begin{equation}
\begin{aligned}
Loss =  \underbrace { -\sum_{x} w(x)(1-p_{l}(x))^{\gamma}g_{l}(x)log(p_{l}(x))}_{\text{Weighted Focal Loss}} \\ -\underbrace{ \frac{2\sum_{x}p_{l}(x)g_{l}(x))}{\sum_{x}p_{l}^{2}(x) + \sum_{x}g_{l}^{2}(x)}}_{\text{Dice Loss}} \label{eq:3}
\end{aligned}
\end{equation} 

where $p_{l}(x)$ is the predicted probability at pixel $x$ to belong to a class $l$, and $g_{l}(x)$ is the pixel ground truth class. For the weighted focal loss, $\gamma$ was set to $2$ and the pixel weight scheme ($w(x)$) proposed by~\cite{quicknat} was used to improve segmentation performance along anatomical boundaries. We additionally included online data augmentation to address two challenges: 1) spatial variations due to head position and image cropping, and 2) intensity inhomogeneities due to scan parameters and movement artefacts (e.g.\ eyes and breathing). The first problem was tackled by applying random spatial transformations (translation, rotation, and global scaling) on the input images.  It is important to notice that spatial augmentations were done in a full image for the segmentation models before cropping, therefore eliminating the intrinsic padding noise when interpolating cropped images. For the second issue, we improved the network's robustness to intensity variations by performing random bias field~\cite{sudre2017longitudinal} and blur transformations. To maintain consistency between neighboring slices, intensity transformations were performed on a subject level (whole volume)  using TorchIO~\cite{perez-garcia_torchio_2020}.

\subsection{MRI Data}\label{sec:mri_data}

MRI scans from the Rhineland Study were collected at two different sites both with identical 3T Siemens MAGNETOM Prisma MRI scanners (Siemens Healthcare, Erlangen, Germany) equipped with 64-channel head-neck coils. The 0.8~$mm$ isotropic T2-weighted 3D Turbo-Spin-Echo (TSE) sequence uses variable flip angles~\cite{busse2008effects} as well as elliptical sampling~\cite{mugler2014optimized} and parallel imaging (PI)~\cite{griswold2002generalized} for faster imaging. For this work, two T2w sequences from the Rhineland Study were considered (referred to as $T2w^{a}$ (original protocol) and $T2w^{b}$). Common sequence parameters are as follows: repetition time (TR) = 2800~$ms$, echo time (TE) = 4405~$ms$, phase-encoding direction: Anterior $>$ Posterior, matrix size = $320 \times 320 \times 224$. The following parameters differ between protocols:  PI acceleration factor: a) 3x1; b) 2x1, PI reference scan: a) integrated; b) external, acquisition time: a) 3:57~$min$; b)~4:47~$min$. Note, care was taken to preserve the image contrast between versions.

 For the training and testing of our pipeline, data from the first 572 participants from the Rhineland Study with a T2w scan was used (referred to as "in-house dataset"). All 572 MRI scans were manually annotated following Section~\ref{sec:labels}. During the creation of the in-house dataset, a group of 12 subjects was separated into another subset (referred to as "no-OB dataset") as these cases were flagged with no visible OB. Subjects without an apparent OB had been reported previously~\cite{weiss2020human}. Consequently, the no-OB cases were used to evaluate the automated method's robustness to an unseen extreme scenario. The remaining sample (n=560) presents a mean age of 53.83 years (range 30 to 87), a mean OB volume of 54.05~$mm^{3}$ (range 12.80 to 111.10~$mm^{3}$), and 56.8\% of subjects are women.  We initially divided the in-house dataset into a training (n=357) and testing (n=203) set. For each subset subjects were randomly selected from sex and age strata to ensure a balanced population distribution. Training data was further split into four groups with the same stratification scheme as before. For a detailed description of the population characteristics of all the aforementioned subsets see Appendix Table~\ref{tab:demograhics}.
 
 Additionally, another subset of the Rhineland Study was selected to evaluate the prediction stability across T2w sequences, as the proposed pipeline was trained only with $T2w^{a}$ scans. As part of the quality assurance workflow in the Rhineland Study before updating a sequence, new incoming subjects are scanned in the same session with both versions for a period of time. After the acquisition reliability is assured the study protocol is updated. Therefore, we selected a group of subjects containing both $T2w^{a}$ and $T2w^{b}$ scans (referred to as "stability dataset", n=109).
 
Finally, we used the publicly available Human Connectome Project (HCP) dataset~\cite{van2012human} to test the generalizability of our method as it contains high-resolutional T2w MR images. A subset of 30 random subjects equally distributed between age categories (22-25, 26-30, and 31-35) was selected. The HCP scans were resampled from isotropic 0.7~$mm$ native resolution to 0.8~$mm$ network input resolution. Finally, manual labels were created for both resolutions using the protocol previously described. HCP data is available at: https://www.humanconnectome.org/study/hcp-young-adult.

\subsection{Evaluation Metrics}
For assessing the segmentation similarity between the predicted label maps and the ground truth, we computed metrics aimed at evaluating different properties: spatial overlap, spatial distance, and volume similarity. We first assessed the spatial overlap as it provides both size and localization consensus by computing the Dice similarity coefficient (Dice), which is a common metric used for validating semantic segmentation performance. Let G (ground truth) and P (prediction) denote binary label maps; the Dice similarity coefficient is mathematically expressed as 

\begin{equation}
\begin{aligned}
         Dice=  \frac{2\cdot \left | G \cap P\right | } { \left | G\right | + \left |  P\right|}
\end{aligned}
\end{equation} where $|G|$ and $|P|$ represent the number of elements in each label map, and $| G \cap P |$ the number of common elements, therefore, the Dice ranges from 0 to 1 and a higher Dice represents a better agreement. However, Dice scores can be drastically affected by small spatial shifts when evaluating small and elongated structures such as the OB~\cite{taha2015metrics,billot2020automated}. Spatial distance-based metrics such as Hausdorff Distance (HD) are widely used for assessing performance in small structures as they evaluate the quality of segmentation boundaries. In this work, we used the Average Hausdorff Distance (AVD), an HD variation less sensitive to outliers. AVD is defined as 
\begin{equation}
\begin{aligned}
AVD(G,P) = \max ( \frac{1}{|G|} \sum_{g \in G} \min_{p \in P} d(g,p),\\ \frac{1}{|P|} \sum_{p \in P} \min_{g \in G} d(p,g) )
\end{aligned}
\end{equation} where $d$ is the Euclidean distance. In contrast to the Dice, AVD is a dissimilarity measurement so a smaller AVD indicates a better boundary delineation with a value of zero being the minimum (perfect alignment). Furthermore, as the OB volumes are usually the desired marker for downstream analysis, we computed a volume-based metric, the volume similarity (VS)~\cite{taha2015metrics}, defined as

\begin{equation}
\begin{aligned}
        VS = 1 - \frac{\left | |G| - |P| \right |}{|G| + |P| }  .
\end{aligned}
\end{equation} While VS is similar to Dice, it does not take into account segmentations overlap and can have its maximum value even when the overlap is zero. In consequence, VS is not used for the localization marker and replaced with localization distance ($R$), a metric more suitable to assess the accuracy of the centroid coordinate created in this stage. Let $p$ and $g$ be the centroid coordinates of the predicted and ground truth label maps, respectively. The localization distance ($R$) is calculated as follows

\begin{equation}
\begin{aligned}
        R(p,q) = \sqrt{(p_{x}-g_{x})^{2} + (p_{y}-g_{y})^{2} + (p_{z}-g_{z})^{2}} .
\end{aligned}
\end{equation} Similar to AVD, a smaller distance indicates improved localization accuracy. Finally, to benchmark performance of various F-CNN  models  we first ranked the models performance for each metric individually and then computed an overall rank as the geometric mean of the model's rankings.  

\section{Experiments and Results}
\begin{table*}[hbt!]
\centering
\caption{Summary of the datasets, number of subjects, T2 protocol and usage for each of the validation experiments}
\resizebox{0.9\textwidth}{!}{
\begin{tabular}{llccc}
\hline
Usage                                   & Dataset Name             & T2 protocol &  Subjects & Cohort\\ \hline
Manual annotation reproducibility  (E1) &  In-house train-set & $T2w^{a}$ & 31  &\multirow{7}{*}{ Rhineland Study}        \\ 
                                        & In-house test-set  &  & 19  &        \\ \cline{1-4} 
Pipeline performance (E2)               & In-house train-set & $T2w^{a}$ & 357  &     \\
                                        & In-house test-set  &  & 203  &      \\ \cline{1-4} 
Age and sex effect sensitivity (E3)     & In-house test-set  & $T2w^{a}$& 203  &       \\ \cline{1-4} 
No apparent OB (E4)                    & No-OB set     & $T2w^{a}$       & 12   &        \\ \cline{1-4} 
Sequence Stability (E5)                 & Stability set &  $T2w^{a}$, $T2w^{b}$   &109            &       \\ \hline
Generalizability (E6)                 & HCP dataset         & $T2w^{hcp}$  &  30  & Human Connectome Project(HCP)       \\ \hline
\end{tabular}
\label{tab:dataset_summary}}
\end{table*} 
In this section, we present six experiments with the aim to thoroughly validate our OB tissue segmentation pipeline. To properly assess the pipeline's performance as a whole, input images to the segmentation stage were pre-processed by the localization stage. Additionally, to ensure that all experiments were carried out under the same testing conditions: All inference analyses were evaluated in a docker container with a 12 GB NVIDIA Titan V GPU (a widely available consumer card). It is important to note, that the pipeline can also run on the CPU.

(E1) We evaluated the OB manual annotations reliability by an inter and intra-rater reproducibility analysis. (E2) We evaluated the performance of each stage of the pipeline against an unseen test-set. We additionally benchmarked the proposed \textit{AttFastSurferCNN} with state-of-the-art F-CNNs and compared the accuracy of one \textit{AttFastSurferCNN} against the proposed ensemble approach of merging four \textit{AttFastSurferCNN} with different training-data conditions. (E3) We assessed the sensitivity of the proposed pipeline to replicate known OB volume effects with respect to age and sex on the test-set against manual labels and benchmark networks. (E4) We evaluated the robustness of the automated method to an extreme and real scenario of cases without an apparent OB. (E5) We tested the stability of the proposed pipeline to variations in acquisition parameters of a T2w sequence. Finally (E6), we accessed the generalizability of our method to different population demographics on the publicly available HCP dataset~\cite{van2012human}. A summary of the data needed for each of the experiments is presented in Table~\ref{tab:dataset_summary}.

\subsection{Manual Annotation Reproducibility (E1)}
To the best of our knowledge, there is no automatic method for detecting and delineating the OB. Therefore, manually annotations are considered the gold standard. As our approach is based on supervised learning, its performance is limited by the quality of the manual annotations. As a consequence, to assess the consistency of the labels created by our main rater, we conducted intra-rater and inter-rater variability experiments.

Fifty random subjects from the in-house dataset were selected. Afterwards cases were manually annotated twice (see Section~\ref{sec:labels}), once by our main rater who had already segmented the cases and once by a second rater trained by our main rater. To remove bias and avoid overestimating performance, raters were blind to the scans' identification; furthermore, the main rater's second segmentations were done with a time gap of two months, and finally, the scans used for training the second rater were not included in the experiment. We assessed intra-rater variability by computing the similarity between the two sets of segmentations of the main rater. Inter-rater variability was estimated by comparing the segmentation agreement between the main rater's first annotations and second rater's annotations.

\begin{figure}[!hbt]
    \centering
    \includegraphics[width=0.48\textwidth]{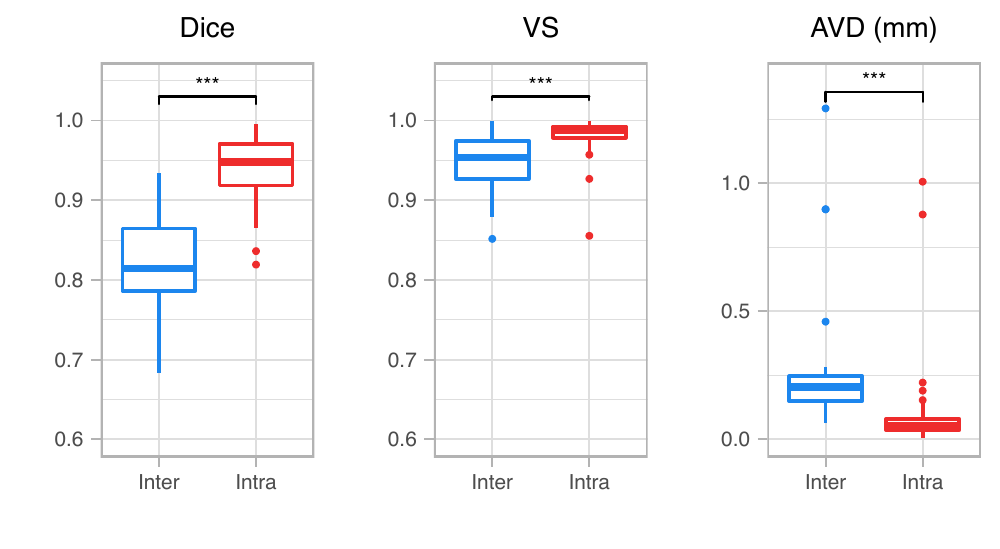}
    \caption{Segmentation similarity scores for total OB comparing intra-rater \textit{vs.}\ inter-rater variability, as well as significance level indicators (paired two-sided Wilcoxon signed-rank). Significance: %$^{\ast}^{\ast}^{\ast}$ p $<0.001$.}
    \textsuperscript{***} p $<0.001$.}
    \label{fig:intra_inter}
\end{figure}

In Figure~\ref{fig:intra_inter}, we present the similarity scores for total OB (left and right combined) in the fifty subjects used for this experiment as well as significance level indicators (paired two-sided Wilcoxon signed-rank test~\cite{wilcoxon1992individual}). We observed that our main rater has a high agreement between labeling sessions (Average~: $Dice=0.9399$, $VS=0.9811$, $AVD=0.0976~mm$). Inter-rater scores (Average~: $Dice=0.8211$, $VS=0.9497$, $AVD=0.2446~mm$) are significantly lower, however, still yield comparable results with other small brain structures inter-rater-scores~\cite{billot2020automated}. These similarity scores put the results of the next section into context where the inter-rater-scores can be seen as the lower-bound of performance and intra-rater-scores as the ideal performance of the automated method.

\subsection{Pipeline Performance (E2)}
In this section, we benchmarked and evaluated the accuracy of each stage of the pipeline in a completely separate unseen test-set. All implemented networks were trained using the scheme mentioned in Section~\ref{sec:model_learning} and data-splits introduced in Section~\ref{sec:mri_data}  were  treated  in  a  leave-one-out  fashion (e.g.\ model 1: splits 2, 3, and 4 were used for training, and split 1 was used for validation).

\subsubsection{ROI Localization}
For evaluating the ability of \textit{FastSurferCNN} to localize the OB ROI in a down-sampled whole-brain image, we trained \textit{FastSurferCNN} from scratch using the four data-splits from the in-house train-set in a leave-one-out cross-validation approach. To ensure good performance and reduce initialization variance, each data-split was trained four times, and the best weights per split were chosen based on the performance in the validation-set. Finally, the model with the highest overall rank of the three evaluation metrics (Dice, AVD, R) in the test-set was selected and incorporated into the pipeline's localization stage. 

We observed that all \textit{FastSurferCNN} models have comparable results when segmenting the ROI (Average~: $Dice\approx 0.83$, $AVD\approx0.4~mm$) with model 4 outperforming models 2 and 3 with statistical significance as illustrated in Appendix Figure~\ref{fig:location_metrics}. However, the small shifts on the predicted label maps did not affect the coordinates from the computed centroid as all models have similar performance ($R\approx 2.08~mm$); hence, any of the trained \textit{FastSurferCNNs} could be used for localizing the ROI for cropping. However, we selected the \textit{FastSurferCNN} trained with data from splits 1, 2, and 3 (model 4) as it has the highest overall rank and outperforms the other versions.

\subsubsection{OB Tissue Segmentation} \label{sec:ob_tissue_segmentation}
To show a proof-of-concept for our proposed \textit{AttFastSurferCNN} in the more difficult task of OB tissue segmentation, we benchmarked our network against state-of-the-art segmentation 2D F-CNNs used for neuro-imaging such as \textit{FastSurferCNN}~\cite{henschel2020fastsurfer}, \textit{UNet}~\cite{ronneberger2015u}, and \textit{QuickNat}~\cite{quicknat}. Additionally, we compared our \textit{AttFastSurferCNN} against 3D networks such as \textit{3D-UNet}~\cite{cciccek20163d} and \textit{3D-FastSurferCNN}, a naive 3D implementation of \textit{FastSurferCNN} by replacing  2D operations for 3D ones. To permit a fair comparison, all benchmark networks followed the same architecture of four encoder blocks, four decoders blocks, and one bottleneck block as  illustrated in Figure~\ref{fig:attFastSurferCNN}. Each block contained the same number of convolutional operations (see Figure~\ref{fig:CDB}) and parameters configuration. All networks were trained in 3 anatomical views (axial, coronal, and sagittal) from scratch with the same training data-scheme; each data-configuration was carried out four different times, and the best weights were selected based on performance in the validation set. Furthermore, the 2D models were implemented with the same multi-slice input method, and 3D models were trained in different anatomical views by permuting the axis from the input volumes just like their 2D counterparts. Finally, all comparative models were implemented with the above-mentioned ensemble approach (see Section~\ref{sec:ensemble}), and segmentation performance on the unseen test set was evaluated by computing three similarity metrics (Dice, AVD, and VS) between the predicted maps and manuals labels.

 \begin{table}[!hbt]
\centering
\caption{Mean (and standard deviation) of segmentation performance metrics of the F-CNN models. Models were ranked ascendingly by individual metrics and the overall rank (geometric mean of the metric rankings). We show significance indicators of the paired Wilcoxon signed-rank test comparing the proposed \textit{AttFastSurferCNN} vs.\ benchmarked F-CNNs. Note \textit{FastSurferCNN} is abbreviated to FSCNN and \textit{AttFastSurferCNN} to AttFSCNN. }
\resizebox{0.48\textwidth}{!}{
\begin{threeparttable}
\begin{tabular}{lccccccc}
\hline
                          & \multicolumn{2}{c}{\textbf{Dice}}    & \multicolumn{2}{c}{\textbf{VS}}                          & \multicolumn{2}{c}{\textbf{AVD (mm)}}                         & \multicolumn{1}{l}{}          \\
                          & Mean (SD) & \multicolumn{1}{l}{Rank} & \multicolumn{1}{l}{Mean (SD)} & \multicolumn{1}{l}{Rank} & \multicolumn{1}{l}{Mean (SD)} & \multicolumn{1}{l}{Rank} & \multicolumn{1}{l}{Overall Rank} \\ \hline
\multirow{2}{*}{AttFSCNN}  & 0.8525    & \multirow{2}{*}{6}       & 0.9104                        & \multirow{2}{*}{6}       & 0.2154                        & \multirow{2}{*}{5}       & \multirow{2}{*}{5.65}         \\
                          & 0.0561    &                          & 0.0634                        &                          & 0.1530                        &                          &                               \\ \hline
\multirow{2}{*}{FSCNN}     & 0.8506   & \multirow{2}{*}{5*}       & 0.9081                        & \multirow{2}{*}{4}       & 0.2134                        & \multirow{2}{*}{6}       & \multirow{2}{*}{4.93}         \\
                          & 0.0577    &                          & 0.0658                        &                          & 0.1488                        &                          &                               \\ \hline

\multirow{2}{*}{QuickNat} & 0.8506   & \multirow{2}{*}{5*}       & 0.9084                       & \multirow{2}{*}{5*}       & 0.2174                       & \multirow{2}{*}{4*}       & \multirow{2}{*}{4.64}         \\
                          & 0.0555    &                          & 0.0635                        &                          & 0.1469                        &                          &                               \\ \hline

\multirow{2}{*}{UNet}      & 0.8473  & \multirow{2}{*}{3**}       & 0.9071                       & \multirow{2}{*}{3*}     & 0.2218                      & \multirow{2}{*}{3**}       & \multirow{2}{*}{3.00}         \\
                          & 0.0610    &                          & 0.0670                        &                          & 0.1567                        &                          &                               \\ \hline
\multirow{2}{*}{FSCNN3D}   & 0.8163  & \multirow{2}{*}{2**}       & 0.8794                      & \multirow{2}{*}{1**}       & 0.2510                      & \multirow{2}{*}{2**}       & \multirow{2}{*}{1.59}         \\
                          & 0.0944    &                          & 0.1109                        &                          & 0.1821                        &                          &                               \\ \hline
\multirow{2}{*}{UNet3D}    & 0.8038  & \multirow{2}{*}{1**}       & 0.8878                      & \multirow{2}{*}{2**}       & 0.2549                      & \multirow{2}{*}{1**}       & \multirow{2}{*}{1.26}         \\
                          & 0.0820    &                          & 0.0950                        &                          & 0.1582                        &                          &                               \\ \hline
\end{tabular}
\begin{tablenotes}  
\item  Significance: \textsuperscript{*} p $< 0.05$, \textsuperscript{**} p $<0.01$
\end{tablenotes}
\end{threeparttable}}
\label{tab:bench_seg_results}

\end{table}

In Table~\ref{tab:bench_seg_results} we present the similarity scores for OB tissue segmentation of all evaluation metrics as well as individual and overall ascending rankings and significance indicators of the two-sided Wilcoxon signed-rank test comparing the proposed \textit{AttFastSurferCNN} vs.\ benchmarked F-CNNs. Here, we observed that our proposed \textit{AttFastSurferCNN} has the highest overall ranking. Additionally,  \textit{AttFastSurferCNN} outperforms all other benchmark networks in all comparative metrics with statistical significance ($p <0.05$) except for \textit{FastSurferCNN}. \textit{FastSurferCNN} outranks our proposed method in AVD, however, there is no statistical difference between them. On the other hand, \textit{AttFastSurferCNN} outperforms \textit{FastSurferCNN} in Dice and VS with a statistical significance ($p <0.05$) in Dice. Finally, it is important to note that all 2D approaches drastically outperform the 3D models with up to 3\% improvement of the Dice, 2.7\% of VS and 4\% of AVD between \textit{UNet} (the lowest rank 2D model) and \textit{3D-FastSurferCNN} (the highest rank 3D model). 

\begin{figure}[!hbt]
    \centering
    \includegraphics[width=0.45\textwidth]{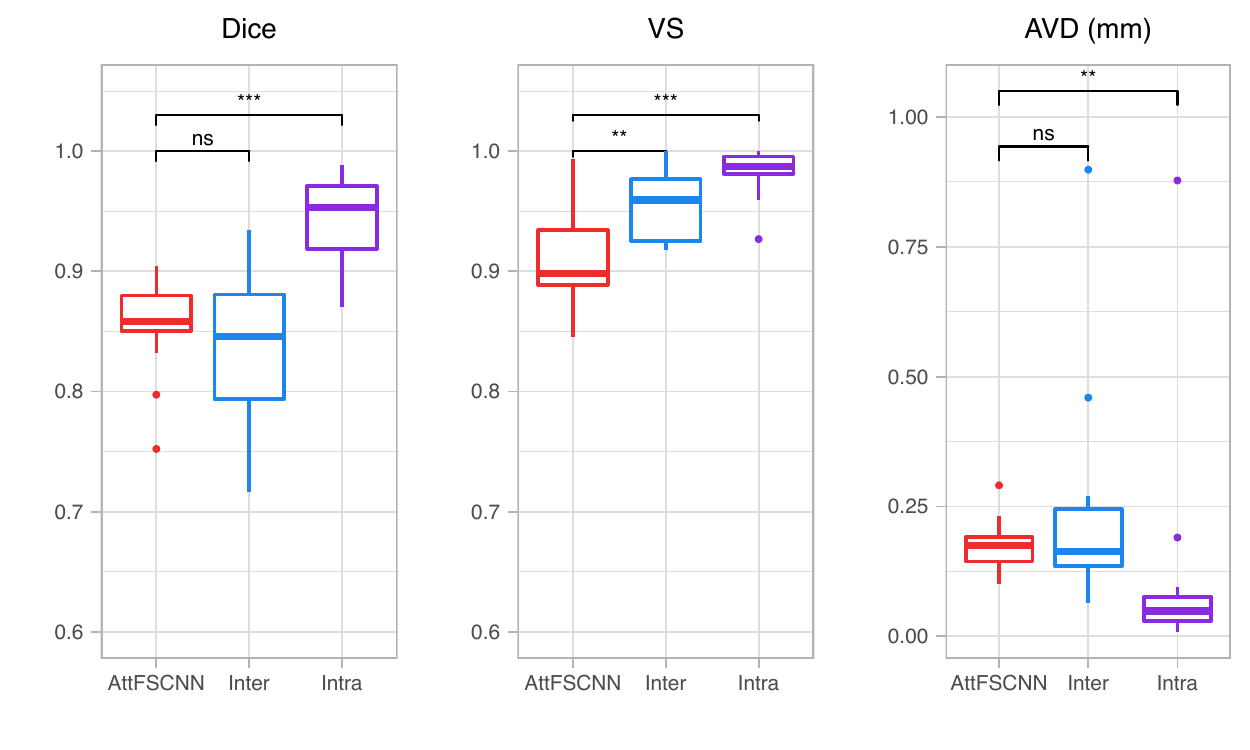}
    \caption{Segmentation similarity scores for total OB comparing \textit{AttFastSurferCNN} (AttFSCNN) \textit{vs.} manual raters (intra- and inter-rater scores), as well as significance level indicators (paired two-sided Wilcoxon signed-rank). Significance: \textsuperscript{***} p $<0.001$   ,\textsuperscript{**} p $<0.01$, ns :  p $\geq0.05$.}
    \label{fig:pipeline_intra_inter}
\end{figure}

Finally, to put the \textit{AttFastSurferCNN} results into context, we compared the performance against the inter and intra-rater variability scores obtained in the manual annotation reproducibility experiment. For a fair comparison, this analysis is exclusively done in 19 cases that are also part of the test-set. Figure~\ref{fig:pipeline_intra_inter} presents box plots for the three accuracy metrics as well as statistical significance indicators (paired two-sided Wilcoxon signed-rank test). We observed that \textit{AttFastSurferCNN} results are significantly lower than the intra-rater scores. However, this was expected as we used the main-rater labels to train our F-CNNs and the intra-rater scores are usually very difficult to reach for an automated method. Moreover, the proposed network outperforms the inter-rater scores (Dice: \textbf{0.8566} vs. 0.8386, and AVD: \textbf{0.1745}~$mm$ vs. 0.2264~$mm$) in localizing the OB tissue and recognizing its boundaries, even if no statistical significance can be inferred from the statistical test. On the other hand, for VS, the inter-rater results are significantly better (VS: 0.9115 vs. \textbf{0.9555}); nevertheless, there is an average VS difference of only $0.04$ between label maps translating to a small volume discrepancy of around 0.020~$mm^{3}$ by every segmented voxel.

\subsubsection{Ensemble}
\begin{figure*}[!hbt]
    \centering
    \includegraphics[width=0.8\textwidth]{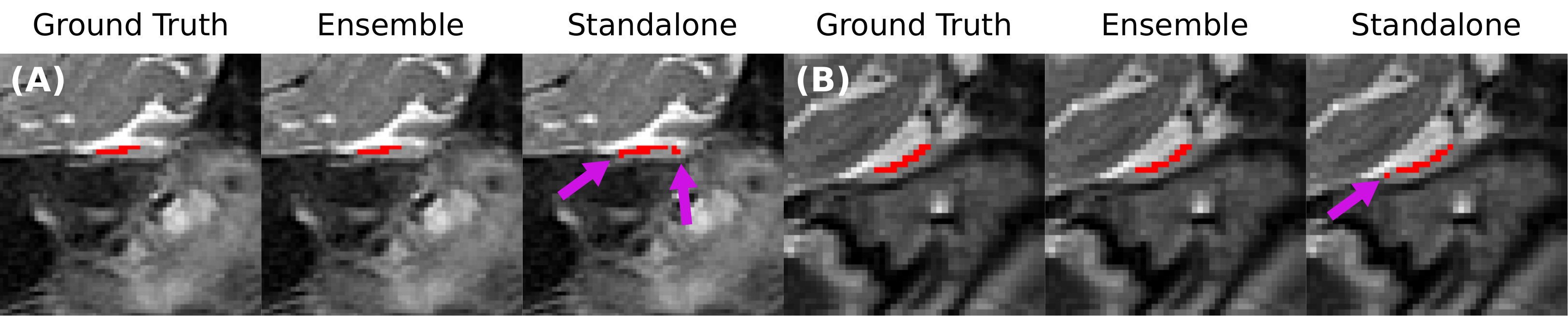}
    \caption{Comparison of the manual ground truth \textit{vs.}\ predictions of the right OB from two subjects on sagittal T2-weighted MRI of the in-house test-set. Purple arrows indicate where the proposed ensemble \textit{AttFastSurferCNN} improves the segmentation over a standalone \textit{AttFastSurferCNN}.}
    \label{fig:ensemblevstandlone}
\end{figure*}

In this section, we tested our ensemble approach of combining the output of four \textit{AttFastSurferCNN} against each individual \textit{AttFastSurferCNN} trained in the previous section. We observed that all standalone models have comparable results in the three similarity metrics (Dice, VS, and AVD) as shown in Table~\ref{tab:ensemble_results}. Thus OB segmentation knowledge is not driven by any particular data-subset, and all \textit{AttFastSurferCNNs} outperform the inter-rater scores for Dice (0.8386) and AVD (0.2264~$mm$). Furthermore, the proposed ensemble model significantly outperforms all standalone (non-ensembled) models with respect to Dice and AVD ($p<0.05$, paired two-sided Wilcoxon signed-rank test). We observed no statistical difference between models in VS except for \textit{AttFastSurferCNN-4}  where the proposed merged method has better results. Finally, we empirically observed that the ensemble model smoothes the label maps slightly, resulting in visually more appealing boundaries as illustrated in Figure~\ref{fig:ensemblevstandlone}.

\begin{table}[!hbt]
\centering
\caption{Mean (and standard deviation) of segmentation performance metrics of the proposed ensemble approach and single \textit{AttFastSurferCNN} (AttFSCNN) models. Models were ranked ascendingly by individual metrics and the overall rank (geometric mean of the metric rankings). We show significance indicators of the paired Wilcoxon signed-rank test comparing the proposed ensemble \textit{AttFastSurferCNN} vs.\ single \textit{AttFastSurferCNN}. }
\resizebox{0.48\textwidth}{!}{
\begin{threeparttable}
\begin{tabular}{lccccccc}
\hline
                                & \multicolumn{2}{c}{\textbf{Dice}} & \multicolumn{2}{c}{\textbf{VS}} & \multicolumn{2}{c}{\textbf{AVD (mm)}} &                       \\
Model                        & Mean(SD)  & Rank                  & Mean(SD)  & Rank                & Mean(SD)  & Rank                 & Overall Rank            \\ \hline
\multirow{2}{*}{Ensemble AttFSCNN} & 0.8525    & \multirow{2}{*}{5}    & 0.9104    & \multirow{2}{*}{3}  & 0.2154    & \multirow{2}{*}{5}   & \multirow{2}{*}{4.22} \\
                                & 0.0561    &                       & 0.0634    &                     & 0.1530    &                      &                       \\ \hline
\multirow{2}{*}{AttFSCNN 3}   & 0.8482    & \multirow{2}{*}{4**}  & 0.9112    & \multirow{2}{*}{4}  & 0.2225    & \multirow{2}{*}{4*}  & \multirow{2}{*}{4.00} \\
                                & 0.0589    &                       & 0.0659    &                     & 0.1706    &                      &                       \\ \hline
\multirow{2}{*}{AttFSCNN 2}   & 0.8477    & \multirow{2}{*}{3**}  & 0.9096    & \multirow{2}{*}{2}  & 0.2234    & \multirow{2}{*}{2**} & \multirow{2}{*}{2.29} \\
                                & 0.0578    &                       & 0.0646    &                     & 0.1614    &                      &                       \\ \hline
\multirow{2}{*}{AttFSCNN 1}   & 0.8476    & \multirow{2}{*}{2**}  & 0.9115    & \multirow{2}{*}{5}  & 0.2276    & \multirow{2}{*}{1**} & \multirow{2}{*}{2.15} \\
                                & 0.0552    &                       & 0.0625    &                     & 0.1749    &                      &                       \\ \hline
\multirow{2}{*}{AttFSCNN 4}   & 0.8469    & \multirow{2}{*}{1**}  & 0.9077    & \multirow{2}{*}{1*} & 0.2230    & \multirow{2}{*}{3**} & \multirow{2}{*}{1.44} \\
                                & 0.0580    &                       & 0.0666    &                     & 0.1491    &                      &                       \\ \hline
\end{tabular}
\begin{tablenotes}  
\item  Significance: \textsuperscript{***} p $<0.001$   ,\textsuperscript{**} p $<0.01$ , \textsuperscript{*} p $< 0.05$
\end{tablenotes}
\end{threeparttable}}
\label{tab:ensemble_results}
\end{table}

\subsection{Age and Sex Effects Sensitivity (E3)}
\begin{table*}[!hbt]
\centering
\caption{ Association of OB volumes (OBV) and age after accounting for sex and head-size (eTIV) on  the in-house test-set  for the  manual  labels (ground truth)  and  benchmark  networks. Linear regression model :  $OBV \sim age + sex + eTIV$. Note \textit{FastSurferCNN} is abbreviated to FSCNN and \textit{AttFastSurferCNN} to AttFSCNN. }
\resizebox{0.8\textwidth}{!}{
\begin{threeparttable}
\begin{tabular}{lccccccc}
\hline
                             & \multicolumn{1}{l}{Ground Truth} & \multicolumn{1}{l}{AttFSCNN} & \multicolumn{1}{l}{FSCNN} & \multicolumn{1}{l}{QuickNat} & \multicolumn{1}{l}{UNet} & \multicolumn{1}{l}{FSCNN3D} & \multicolumn{1}{l}{UNet3D} \\ \hline
\multirow{2}{*}{(Intercept)} & 53.292***                  & 55.517***                    & 54.774***                 & 56.038***                     & 55.330***                & 45.714***                  & 47.186***                   \\
                             & (1.953)                    & (1.636)                     & (1.620)                   & (1.642)                       & (1.638)                  & (1.535)                    & (1.501)                     \\ \hline
\multirow{2}{*}{Age}         & -0.319***                  & -0.232**                     & -0.204**                  & -0.213**                      & -0.211**                 & -0.225**                  & -0.241***                    \\
                             & (0.092)                    & (0.077)                      & (0.076)                   & (0.077)                       & (0.077)                  & (0.072)                    & (0.070)                     \\ \hline
\multirow{2}{*}{Sex: m/f}    & 5.940                      & 3.150                        & 2.612                     & 3.409                         & 3.017                    & 1.980                      & 2.897                       \\
                             & (3.463)                    & (2.900)                      & (2.871)                   & (2.910)                       & (2.903)                  & (2.721)                    & (2.660)                     \\ \hline
\multirow{2}{*}{eTIV}        & 14.286                     & 32.189***                    & 32.297***                 & 31.713***                     & 32.590***                & 25.022**                   & 21.116**                    \\
                             & (10.238)                   & (8.577)                      & (8.490)                   & (8.605)                       & (8.586)                  & (8.047)                    & (7.867)                     \\ \hline
R-squared                    & 0.124                      & 0.205                        & 0.193                     & 0.199                         & 0.199                    & 0.157                      & 0.156                       \\ \hline
N                            & 203                        & 203                          & 203                       & 203                           & 0.203                    & 0.203                      & 203                         \\ \hline
\end{tabular}
\begin{tablenotes}  
\item  Significance: \textsuperscript{***} p $<0.001$   ,\textsuperscript{**} p $<0.01$ , \textsuperscript{*} p $< 0.05$
\end{tablenotes}
\end{threeparttable}}
\label{tab:lineal_regresions}
\end{table*}
OB volumes obtained from manual segmentations of T2w images have shown to be negatively correlated with age~\cite{hummel2011correlation,buschhuter2008correlation,hummel2015volume}. Therefore, any automated method that intends to detect this small structure should be able to replicate these effects. As a consequence, we evaluated the sensitivity of our proposed pipeline to replicate  ground truth age dependencies in the in-house unseen test-set ($n=203$) which has a comparable size to other manually annotated OB datasets~\cite{buschhuter2008correlation,hummel2015volume} used for volume correlations. Furthermore, we compared our results with the F-CNNs used in the benchmark (see Section~\ref{sec:ob_tissue_segmentation}). The association of OB volumes (OBV) and age was assessed using a linear regression after accounting for sex and head-size (estimated total intracranial volume, eTIV) (Model: $OBV \sim age + sex + eTIV$). All statistical analyses were performed in R~\cite{R} and eTIV estimations were computed using \textit{FreeSurfer}~\cite{freesurfer1,freesurfer2,buckner2004unified}.

All predicted OB volumes significantly decreased with age as can be seen in Table~\ref{tab:lineal_regresions}, which in turn follows the behavior of the manual data and other studies~\cite{hummel2011correlation,buschhuter2008correlation,hummel2015volume}. We found an improvement in the modeling ($R^{2}$) of the age effects in the \textit{AttFastSurferCNN} compared to the ground truth and the other comparative networks. Finally, we did not find a sex difference for any of the models, and, as expected, the inferred OBV are positively associated with eTIV (see Table~\ref{tab:lineal_regresions}).

\subsection{E4: No Apparent Olfactory Bulb (E4)}
As the proposed pipeline is to be deployed as a post-processing OB analysis pipeline for the T2w MRI of the Rhineland Study, it should be robust to cases without an apparent OB that - based on the size of our in-house dataset - occur with  an approximate prevalence of 2\%. In this section, we processed the 12 flagged cases with no apparent OB and evaluated the OB volume estimates. Note, all cases used for training our \textit{AttFastSurferCNN} have a visible OB.

\begin{figure}[!hbt]
    \centering
    \includegraphics[width=0.45\textwidth]{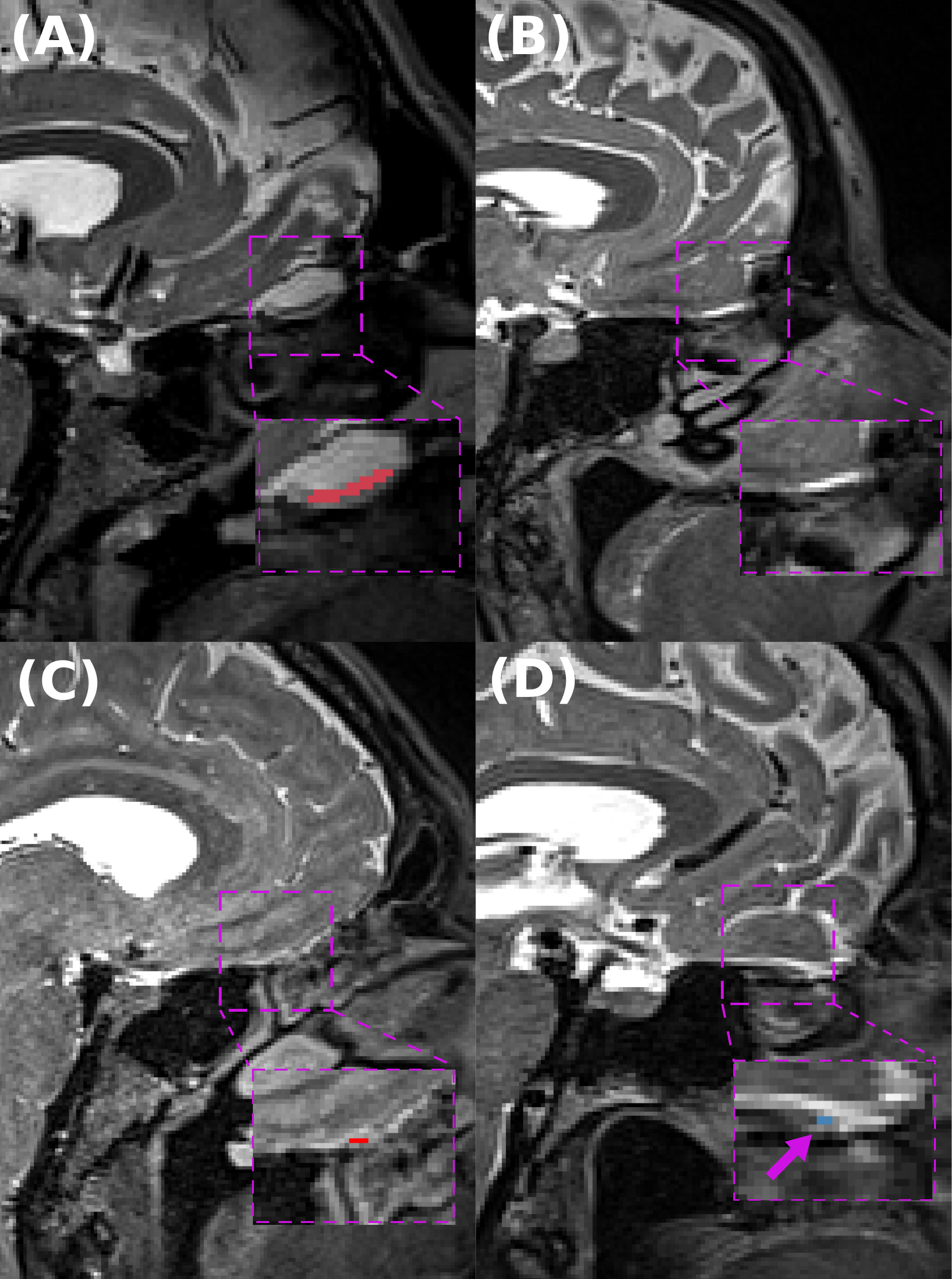}
    \caption{ A-D) Sagittal T2-weighted MR images and predictions on cases from the Rhineland Study. A) Normal subject from the in-house dataset with a visible OB, B) Subject without an apparent OB where the pipeline also agrees with our main rater. C-D) Subjects flagged with no visible OB by our main rater, however, the pipeline still predicts some voxels as OB (total volume $<10.2~mm^{3}$) due to the current resolution our raters cannot reliably assess the predicted segmentation. Note, red indicates Right OB and blue left OB (purple arrow indicate the segmented voxel).}
    \label{fig:no_ob_fig}
\end{figure}

The automated method agreed with the main-rater in 50\% percent of these cases as illustrated in Figure~\ref{fig:no_ob_fig} B) and shown in Appendix Figure~\ref{fig:no_ob_vol}. For the remaining cases: three had a total predicted volume smaller than 2.5~$mm^{3}$ and the other three between 7~$mm^{3}$ to 10.2~$mm^{3}$. We additionally observed that there is hemisphere asymmetry where the maximum predicted volume by any hemisphere was 8.7~$mm^{3}$ translating in a detection of only 17 voxels. After visually inspecting the predicted label maps by two different raters, we observed that with the current resolution our raters cannot reliably assess the predicted segmentation of an individual olfactory bulb with a size smaller than 10~$mm^{3}$ as seen in Figure~\ref{fig:no_ob_fig} C) and D) where the in-plane segmentation is only a few voxels. For this reason, we additionally evaluated the effects of OB size on the segmentation accuracy of the automated method for the test-set. We observed that segmentation performance decreases in subjects with a total OB smaller than 20~$mm^{3}$. Furthermore, OB volumes are positive correlated with  similarity metrics (Dice: $R=0.39,p<0.001$, VS: $R=0.23,p<0.001$) and negative correlated with AVD ($R=-0.39,p<0.001$), a dissimilarity metric. For more detailed information see Appendix Figure~\ref{fig:vol_vs_metric}.

\subsection{Sequence Stability (E5)}
In this section, we processed all  $T2w^{a}$ and $T2w^{b}$  scans from the stability dataset with the proposed pipeline. Afterwards, we assessed the pipeline stability by comparing the similarity of total OB volume across sequences by volume similarity (VS) as described in the metric evaluation section. Additionally, we calculated the agreement of total OB volume between sequences by an intra-class correlation (ICC) using a two-way fixed, absolute agreement and single measures with a 95\% confidence interval (ICC(A,1))~\cite{mcgraw1996forming}. To further compare the agreement between sequences, three random subjects from the stability dataset were selected and both T2w sequences were manually annotated. Subsequently, segmentation performance metrics (Dice, VS, AVD) between the manual and predicted label maps were computed. It is important to note that we did not compute overlap segmentation performance metrics (Dice and AVD) across different sequence label maps of the same subject as this would require registering the scans. It would not only include inherent variance from acquisition noise (e.g.\ motion artefacts, non-linearities based on different positioning) but also variance due to registration inaccuracies and interpolation artefacts. 

After visual quality inspection, a total of 7 scans were excluded from this analysis due to image artefacts such as motion or low contrast (see Appendix Figure~\ref{fig:motion} for two examples). For the remaining cases (n=102), we observed a good agreement between the $T2w^{a}$ and $T2w^{b}$ sequences (ICC: 0.897 {[}0.845 - 0.931{]}) and a volume similarity (VS: 0.889 (0.090)) comparable to the one described in previous sections. However, we observed a statistical difference between volume estimates ($p<0.01$, paired two-sided Wilcoxon signed-rank test). Furthermore, to give more context on how variations in a T2w sequence affect the pipeline's predictions, we analyzed the segmentation similarity on the manually annotated subset. As expected, the result on the $T2w^{a}$ (training) sequence outperforms the  $T2w^{b}$ segmentation results (Dice: \textbf{0.8622} vs.\ 0.8597, VS: \textbf{0.9343} vs.\ 0.9066 and AVD: \textbf{0.1816}~$mm$ vs.\ 0.1965~$mm$). Nevertheless, the segmentation performance in both sequences is in the range of intra-rater scores (Dice: 0.8386, VS: 0.9555, and AVD: 0.2264~$mm$). Demonstrating that systematic sequence improvements can be beneficial in an ongoing population study without diminishing the performance of the proposed method. Even though our pipeline showed volume stability across sequences and that segmentation performance was not affected, it is still important to control for MRI sequence in any downstream statistical analysis when including data from multiple MRI sequences.

\begin{figure*}[!hbt]
    \centering
    \includegraphics[width=0.9\textwidth]{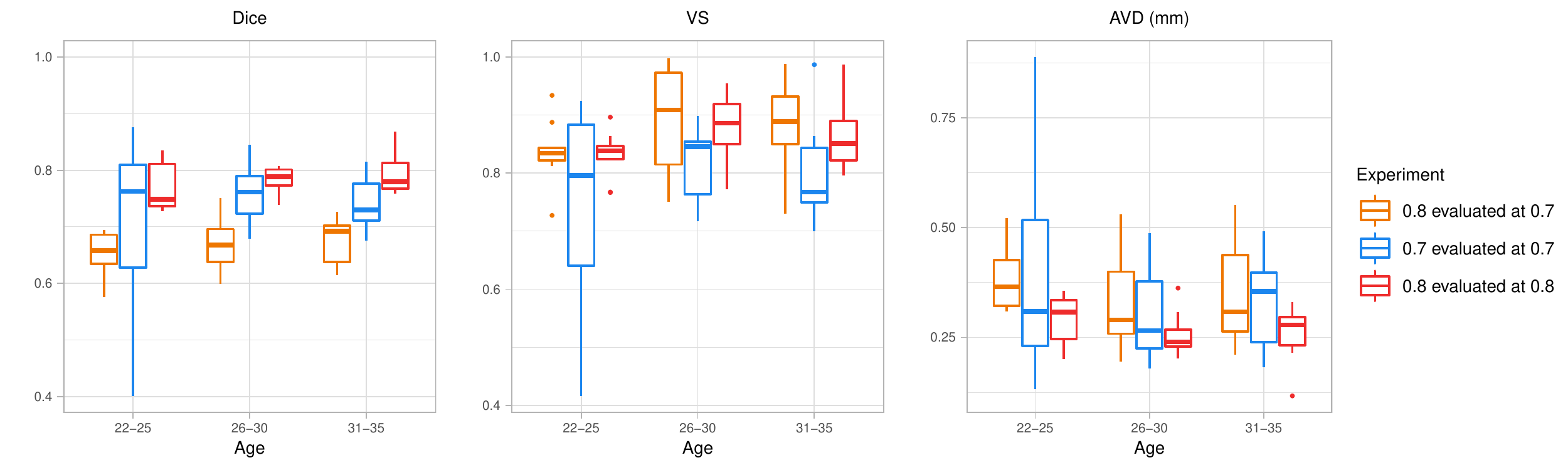}
    \caption{Segmentation similarity scores of total OB  for the 30 labelled cases from the HCP dataset stratified by age category, as well as comparison of the pipeline's performance at native HCP resolution (0.7~$mm$ isotropic, with upsampling: orange, directly: blue) and at the networks original training resolution (0.8~$mm$ isotropic, red).}
    \label{fig:hpc_metrics}
\end{figure*}

\subsection{Generalizability (E6)}
The lack of MR hardware heterogeneity (i.e.\ scanners, field strength, and acquisition parameters) in our training set can limit the ability of the neural network to generalize to unseen T2w images acquired under different conditions. In order to quantify the robustness of our pipeline, we tested it on 30 subjects of the HCP dataset, acquired with a different resolution with isotropic 0.7~$mm$ voxels. In addition to sequence differences, HCP images are de-faced. In order to analyze our method at the native 0.7~$mm$ HCP resolution as well as at the default 0.8~$mm$ network resolution, we constructed manual annotations twice per subject, one for each resolution. 

We perform three experiments: 
A) Input images were resampled to the default network resolution (isotropic 0.8~$mm$), resulting label maps were upsampled to the original 0.7~$mm$ resolution and compared to the manual reference there.
B) Images were processed directly at the native resolution of 0.7~$mm$ and compared to the 0.7~$mm$ manual reference, thus, evaluating the networks' generalizability to segment inputs at a slightly higher and unseen resolution directly. 
C) Same as A) but instead of upsampling final labels they are compared with the manual reference delineated at 0.8~$mm$, avoiding the final upsampling step. This permits quantifying the accuracy for the default behaviour of the network, if final segmentations at 0.8~$mm$ are sufficient for the user. 

Figure~\ref{fig:hpc_metrics} clearly indicates that option A (orange) provides the lowest performance, most likely due to the fact that it includes interpolation artefacts from upsampling the final labels. Resampling label maps is often problematic and should be avoided. If final results are required at the original (here 0.7~$mm$) resolution it is indeed better to directly segment these images at the native resolution (option B, blue boxes). Even though the network has not been trained on this resolution, it can generalize remarkably well. Option C demonstrates that best results can be obtained at the default network resolution of 0.8~$mm$, which is the recommended approach. 

\begin{figure*}[!hbt]
    \centering
    \includegraphics[width=\textwidth]{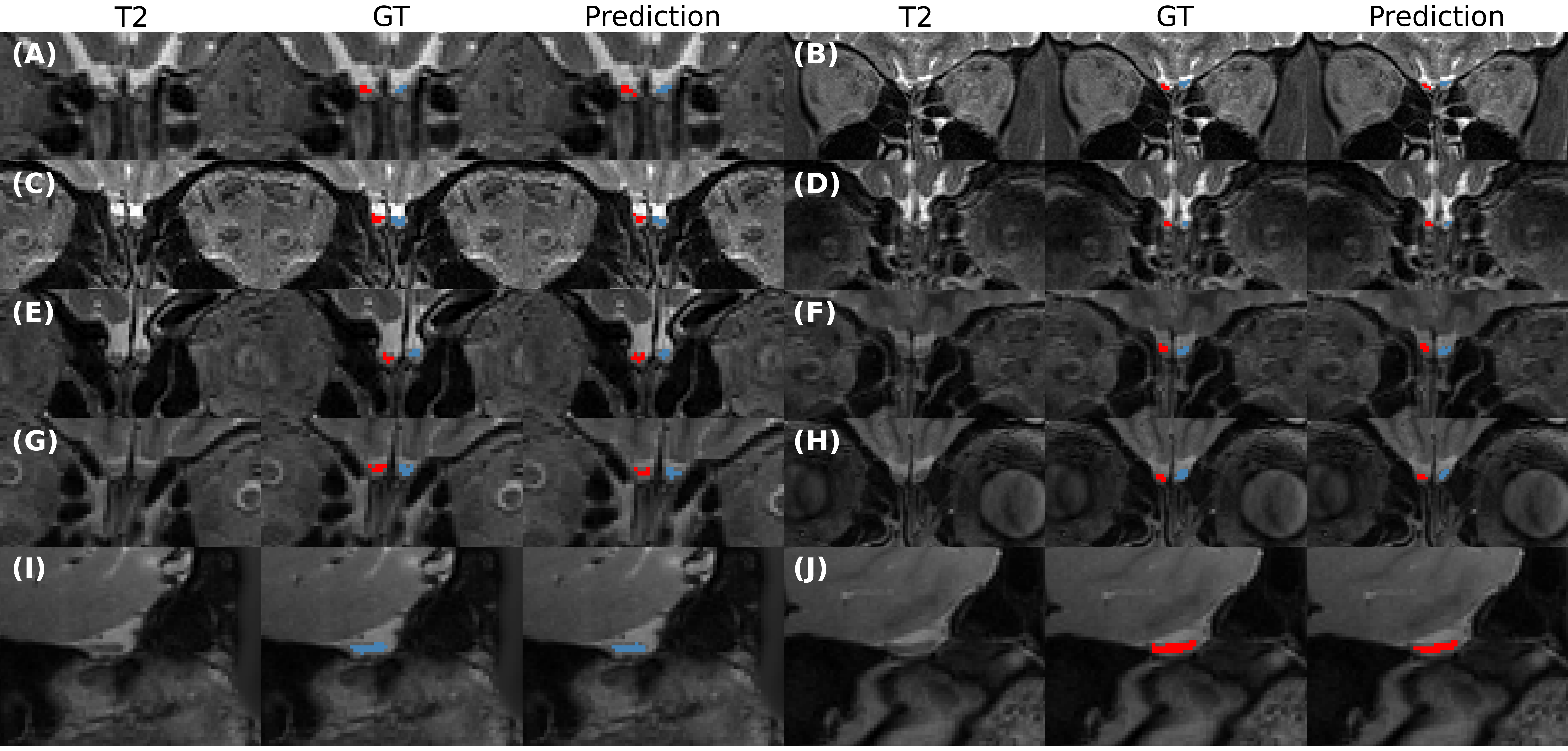}
    \caption{Comparison of the ground truth \textit{vs.}\ predictions on coronal (A-H) and sagittal (I-J) T2w MRI from subjects of the Rhineland Study (A-E) and HCP (F-J) dataset at 0.8~$mm$. A-J) Accurate automatic segmentation of total OB on a heterogeneous population. Note, blue: left OB and red: right OB.}
    \label{fig:outcome_examples}
\end{figure*}

 As expected, overall performance on HCP data is slightly lower than the results obtained on our in-house dataset (see Section~\ref{sec:ob_tissue_segmentation}). The HCP dataset, however, consists of de-faced scans (never encountered during training) from a younger age distribution, and was acquired with different acquisition parameters. Due to these differences, segmentation scores are not directly comparable. Nevertheless, the proposed pipeline generalizes quite well across age-categories, especially when evaluated at the original training resolution as metrics remained relatively stable with an overall good performance (Dice: 0.7816, VS: 0.8583, and AVD: 0.2683~$mm$, red boxes). Additionally, we observe that segmentation accuracy decreases slightly for ages outside the training range (namely 22 to 25, training data started at age 30). Yet the overall high accuracy shows that our proposed pipeline can robustly generalize to the unseen HCP data. Examples of OB segmentations for both the in-house as well as the HCP dataset can be found in Figure~\ref{fig:outcome_examples}. 

\section{Discussion}
In this work, we established, validated, and implemented a novel deep learning pipeline to segment and quantify the olfactory bulb on high resolutional T2-weighted MR scans. The proposed pipeline is fully automatic and can analyze a 3D volume in less than a minute in an end-to-end fashion, even though it implements a three-stage design. The use of deep learning components for localizing and segmenting the OB enables the pipeline to accurately and quickly quantify the OB volume, providing a robust and reliable solution for assessing  OB volumes in a large cohort study.

Segmenting the OB in T2w scans is a challenging task due to size, sensitivity to artefacts, age effects, and visibility on MR images (partial volume effects). Despite all these challenges, we demonstrate the feasibility of segmenting the OB on high resolutional isotropic T2w MR images. Our main rater's manual annotations exhibit a high intra-rater reliability in terms of boundary delineation, OB localization, and volume estimation. Furthermore, we verified the reproducibility of our labeling protocol with inter-rater reliability similar to the one reported in other manually annotated medical datasets~\cite{billot2020automated,estrada2020fatsegnet}. We cannot directly compare the segmentation performance with other studies that manually labeled the OB on T2w MR images as they only report the volume difference for repeated measurements by a single observer or across observers~\cite{hummel2011correlation,obsemi,mueller2005olfactory,yousem1997reproducibility}. Nonetheless, the volume similarity for both inter and intra-rater variability yields comparable or even better results  than the OB studies mentioned above. These results demonstrate the quality of the manual annotations and soundness of developing an automated method for segmenting the OB using a supervised learning technique.

For the first stage of the pipeline, i.e.\ localization of the OB in a whole-brain image, all four implemented \textit{FastSurferCNNs} can successfully localize a forebrain region containing the OBs from both hemispheres (region of interest) and determine a cropping coordinate based on the centroid from a segmentation prediction map. However, for our final localization model, we chose the \textit{FastSurferCNN} model 4 as it outranked all other models in all evaluation metrics (Dice, VS, and R). The implemented localization block is able to identify the region of interest in a low-resolution image even when the input scans are defaced as in the HCP dataset or have motion artefacts as illustrated in Appendix Figure~\ref{fig:motion}. 

For the more challenging task of segmenting OB, we contribute a deep learning architecture (\textit{AttFastSurferCNN}) by incorporating a self-attention module inside our \textit{FastSurferCNN}. The introduction of a self-attention mechanism improves the network's modeling of global dependencies~\cite{fu2019dual,zhang2019self}, thus increasing the attention to spatial information and boosting the learning of such a fine-grained structure as the OB. We demonstrate that \textit{AttFastSurferCNN} recovers OB significantly better than the standard \textit{FastSurferCNN} and other traditional deep learning variants used for semantic segmentation.  It is also important to note that our proposed method shows an improvement when evaluating volume associations in a large cohort despite the slight changes at the image metric level.
Additionally, each of the four individual \textit{AttFastSurferCNN}s that compos the ensemble model outperforms manual inter-rater scores for segmenting and delineating the OB. Even though the volume similarity from the proposed method is lower than the one from the manual raters, the mean volume difference ($\approx 9\%$) is still in the 10\% acceptable difference 
used as selection criteria by other studies for including the OB volumes of a subject with multiple manual annotations~\cite{obsemi}. Moreover, the implemented assemble approach regularizes the predicted segmentation by combining the spatial context from different views and models, ultimately improving the segmentation of the OB boundaries and reducing the variance due to networks initialization. Furthermore, the predicted probability maps from all individual \textit{AttFastSurferCNNs} can be used to compute the pipeline uncertainty~\cite{kendall2017uncertainties,roy2019bayesian}, a potential quality control marker for flagging problematic cases.

The 2.5D approach used for all 2D benchmark networks of multi-network view-aggregation and multi-slice input drastically outperforms the comparative 3D models. Showing that 3D methods are not always the best method and that 2D models can yield better results when strategies to increase the spatial information are included as the one used in this work. Moreover, reducing the scope of the local neighbourhood when segmenting a small structure like the OB is beneficial as it reduces the amount of redundant information and increases the attention to the spatial information surrounding the OB. To improve attention in a 3D network towards OB, a naive solution would be to include the proposed self-attention layer. However, the computation of an attention map of size $NxN$, where $N$ are the number of voxels, will considerably increase the GPU memory requirements and 3D networks are inherently memory expensive to train. Therefore, a self-attention layer is not an efficient and scalable solution for this type of networks. More efficient techniques are being studied, but they are outside the scope of this paper.

As demonstrated in the Rhineland data, the proposed pipeline successfully identifies the OB on a T2w scan as seen in Figure~\ref{fig:outcome_examples} A) to E). The pipeline also replicates the negative correlation of OB volumes with age reported in previous studies~\cite{hummel2011correlation,buschhuter2008correlation,hummel2015volume} and also visible in our manual annotations. We, furthermore, detected no sex difference after accounting for head size, however, estimates from \textit{AttFastSurferCNN} and all comparative networks are positively correlated with head size - a result that is also detected in the manual segmentations - as expected - but with a lower significance and magnitude. All automated methods show stronger and less variable eTIV effect across subjects (see Table~\ref{tab:lineal_regresions}), explaining the significance discrepancy. The difference in effect magnitudes can be attributed to the F-CNN's ability to learn consistent information across subjects exhibiting stability to random noise and thus generating smoother segmentations than manual raters. Furthermore, our proposed pipeline efficiently handles cases without an apparent OB by not segmenting the structure at all or only a few voxels ($<10~mm^{3}$) as seen in Figure~\ref{fig:no_ob_fig}~B), C), and D). Additionally, the sequence stability dataset demonstrates a good agreement of volume estimates between sequences. It must be noted that the difference in volume estimates includes not only potential variances of the processing pipelines but also variance from acquisition noise (e.g.\ motion artefacts, non-linearities based on different head positions). Therefore we recommend controlling for MRI sequence in follow-up statistical analysis when pooling input data. As consistent changes in a sequence can reflect a consistent change in measured OB size. Nonetheless, segmentation performance in all sequences yields comparable results to the manual inter-rater scores.  The fact that our results in the Rhineland Study data (i) replicate known OB volume effects, (ii) properly identify scans without an apparent OB, and (iii) demonstrate a good agreement of volume estimates among variations of the study's T2w sequence corroborates robustness and stability of our pipeline. Nevertheless, due to current image resolution and based on quality assessment of all the predicted label maps generated in this work, we recommend visually inspect cases with an OB volume smaller than 20~$mm^{3}$ before including them in any analysis.

Our automated method not only exhibits generalizability across a wide range of ages from the Rhineland Study but can also extend to another population distribution (HCP dataset) with different acquisition parameters. The pipeline presents optimal results when the input images have the default training resolution of 0.8~$mm$ isotropic. Nonetheless, results at a different resolution (HCP native resolution of 0.7~$mm$) still yield a good performance even with all the various other differences, e.g.\ different distribution, de-faced image, acquisition parameters, and image resolution. Even though our method shows robustness to de-facing pre-processing steps in HCP, de-facing or skull stripping can be problematic due to the proximity of the OB region to the cropped mask, in the worst case scenario - depending on the method - resulting in accidentally cropping into the OB. Therefore, full head T2w scans are the recommended input to our pipeline. Additionally, T2w scans with a different resolution from the ones presented in this work can also be analyzed by running the pipeline with the default behaviour (resampling inputs to 0.8~$mm$) or by processing inputs directly at the native image resolution, if it is close to 0.8~$mm$ isotropic. In these cases is highly recommended, however, that segmentation quality is assessed by the user.
Generally, since the pipeline is based on deep learning, the model can easily be fine-tuned to another desired resolution by retraining or by more aggressive scaling augmentation techniques. 

In conclusion, we have developed a fully automated post-processing pipeline for OB segmentation on sub-millimeter T2-weighted MRI based on advanced deep learning methods. To the best of our knowledge, the presented pipeline is the first to accurately segment the OB in a large cohort and is meticulously validated not only against segmentation accuracy but also with respect to known OB volume effects (e.g.\ age). 

\section{Acknowledgment}
We would like to thank the Rhineland Study group for supporting the data acquisition and management. This work was supported by DZNE institutional funds, the Federal Ministry of Education and Research (BMBF), Germany (grant numbers: 031L0206 and 01GQ1801), the Diet-Body-Brain Competence Cluster Nutrition Research funded by the BMBF (grant numbers: 01EA1410C and 01EA1809C), and the NIH (grant numbers: R01NS083534, R01LM012719, P41EB01589620). Ran Lu was partially supported by a scholarship from China Scholarship Council.

External data used in the preparation of this work were obtained in part by the Human Connectome Project, WU-Minn Consortium (Principal Investigators: David Van Essen and Kamil Ugurbil; 1U54MH091657) funded by the 16 NIH Institutes and Centers that support the NIH Blueprint for Neuroscience Research; and by the McDonnell Center for Systems Neuroscience at Washington University.

\bibliographystyle{elsarticle-num}
\bibliography{mybibfile.bib}

\newpage
\onecolumn
\section*{Appendix}
\setcounter{table}{0}
\setcounter{figure}{0}

\begin{table*}[hbt]
    \centering
        \caption{OB demographics for the total in-house dataset and for the training and testing subsets. Descriptive data were expressed as mean (SD) or count (percentage) for continuous or categorical variables, respectively. Inter group differences were compared with the Student’s t-test for continuous variables and with the Pearson’s chi-square test for categorical variables.}
    \label{tab:demograhics}
    \resizebox{0.9\textwidth}{!}{
        \begin{tabular}{l c c c c c c r}
         & \multicolumn{4}{c}{Trainset} & \multirow{2}{*}{Testset (N=203)} & \multirow{2}{*}{Total (N=560)} & \multirow{2}{*}{p value}\\ 

         & Split\_1 (N=90) & Split\_2 (N=89) & Split\_3 (N=89) & Split\_4 (N=89) &  &  &  \\
        \hline
        Sex &  &  &  &  &  &  & 0.996\\
        \hline
        \ Female & 52 (57.8\%) & 52 (58.4\%) & 50 (56.2\%) & 50 (56.2\%) & 114 (56.2\%) & 318 (56.8\%) & \\
        \hline
        \ Male & 38 (42.2\%) & 37 (41.6\%) & 39 (43.8\%) & 39 (43.8\%) & 89 (43.8\%) & 242 (43.2\%) & \\
        \hline
         Age &  &  &  &  &  &  & 0.992\\
        \hline
        \ Mean (SD) & 53.900 (12.986) & 53.360 (13.487) & 53.708 (13.345) & 54.348 (12.763) & 53.837 (13.540) & 53.832 (13.247) & \\
        \hline
        \ Range & 30.000 - 81.000 & 31.000 - 85.000 & 30.000 - 82.000 & 31.000 - 83.000 & 30.000 - 87.000 & 30.000 - 87.000 & \\
        \hline
        OB Volume($mm^{3}$) &  &  &  &  &  &  & 0.126\\
        \hline
        \ Mean (SD) & 52.173 (14.623) & 53.064 (15.814) & 55.342 (13.353) & 51.424 (13.629) & 55.896 (18.576) & 54.049 (16.085) & \\
        \hline
      \ Range & 19.456 - 84.480 & 24.576 - 88.064 & 29.696 - 84.992 & 21.504 - 84.480 & 12.800 - 111.104 & 12.800 - 111.104 & \\
        \hline
        \end{tabular}}
\end{table*}

\begin{figure*}[!hbt]
    \centering
    \includegraphics[width=0.7\textwidth]{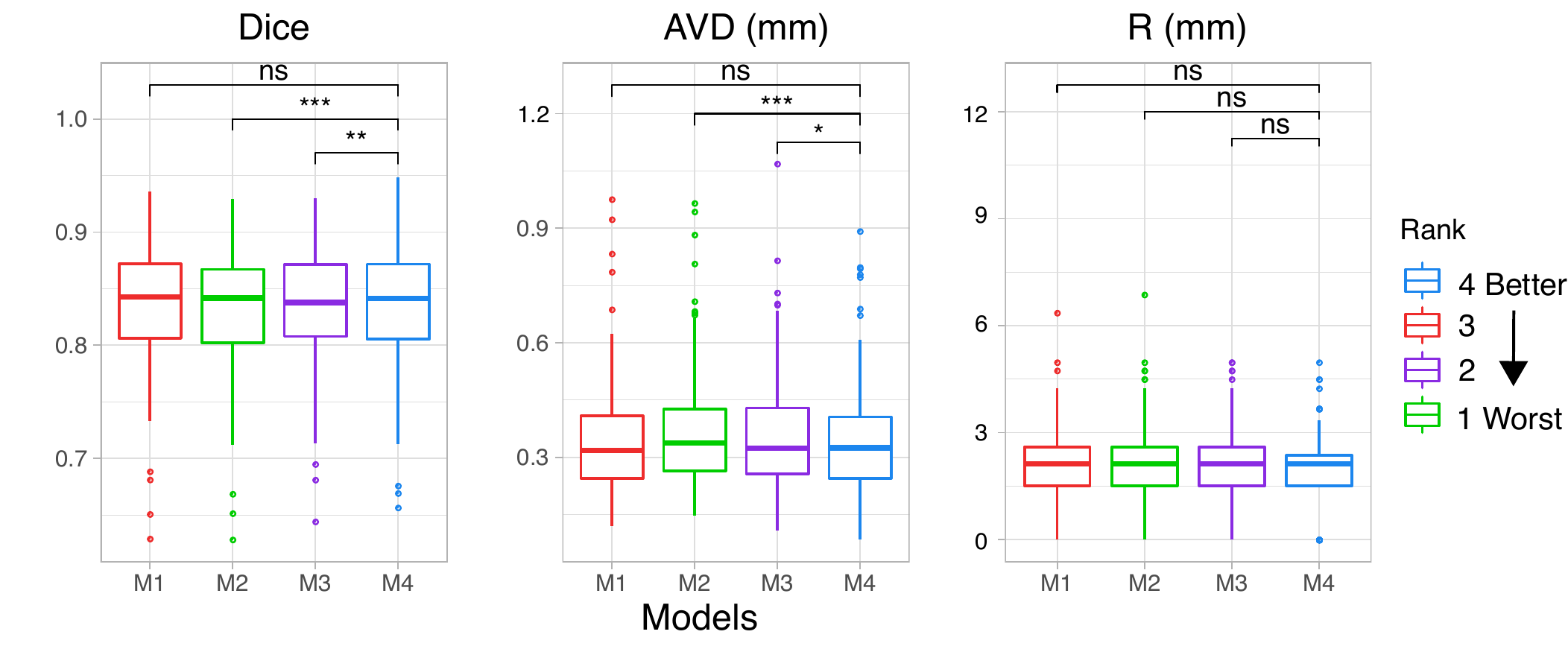}
    \caption{Similarity metrics scores for ROI localization comparing all trained \textit{FastSurferCNN} models. Models were ranked ascendingly by individual metrics (box-plot color) and the overall rank (geometric mean of the metric rankings). We show significance level indicators of the paired Wilconox signed-rank test comparing \textit{FastSurferCNN-4} (M4, model with best overall rank) against the other \textit{FastSurferCNNs} (M1,M2,M3). Significance: \textsuperscript{***} p $<0.001$   ,\textsuperscript{**} p $<0.01$ , \textsuperscript{*} p $< 0.05$, ns : p $\geq 0.05$.}
    \label{fig:location_metrics}
\end{figure*}

\begin{figure*}[!hbt]
    \centering
    \includegraphics[width=0.5\textwidth]{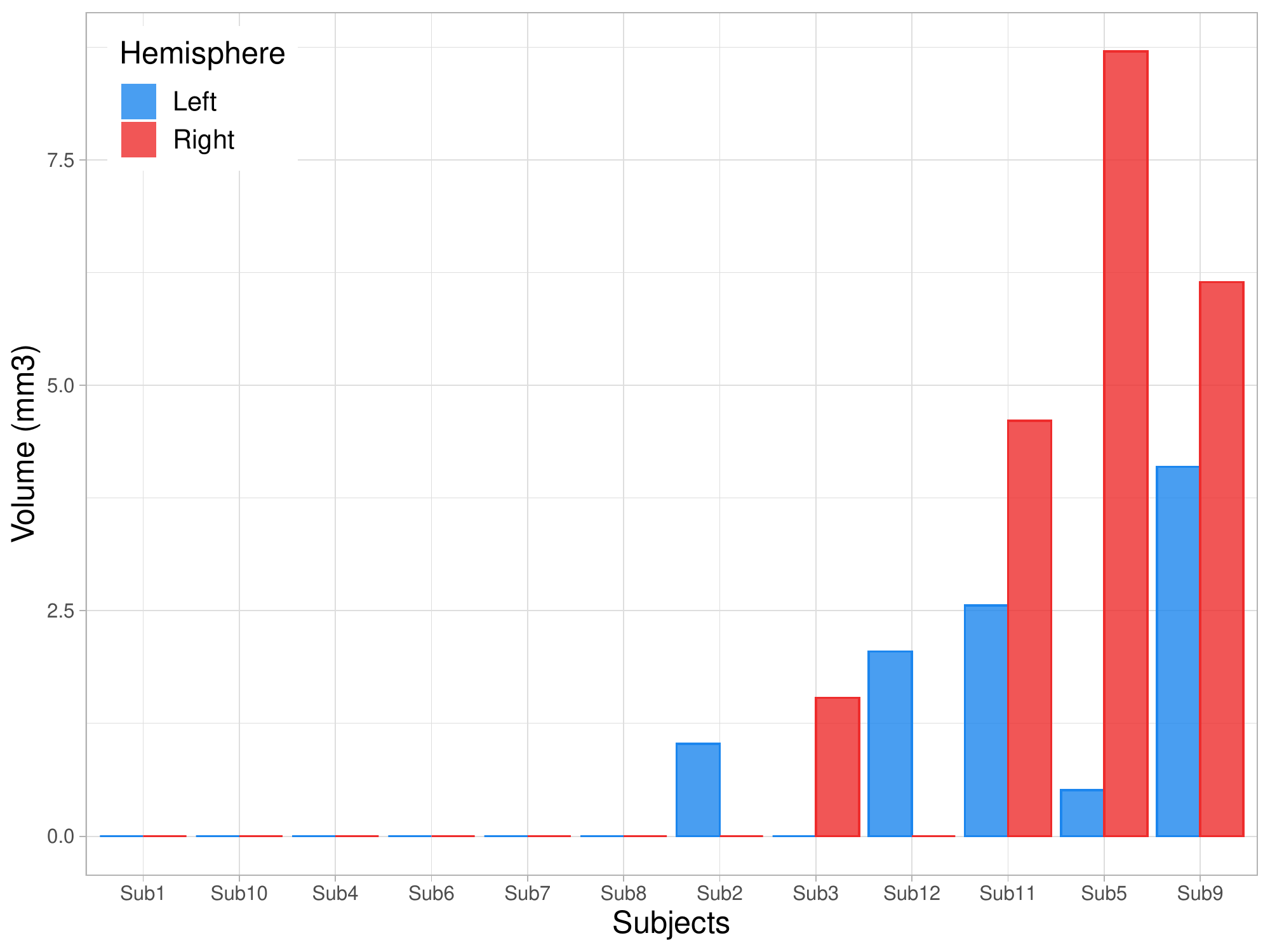}
    \caption{OB volume estimates (left (blue) and right (red)) after processing the 12 subjects flagged with no visible OB with the proposed automated pipeline. The automated method agreed with the main-rater in 50\% percent of these cases. For the remaining cases: three had a total predicted volume smaller than 2.5~$mm^{3}$ and the other three between 7~$mm^{3}$ to 10.2~$mm^{3}$.}
    \label{fig:no_ob_vol}
\end{figure*}

\begin{figure*}[!hbt]
    \centering
    \includegraphics[width=0.9\textwidth]{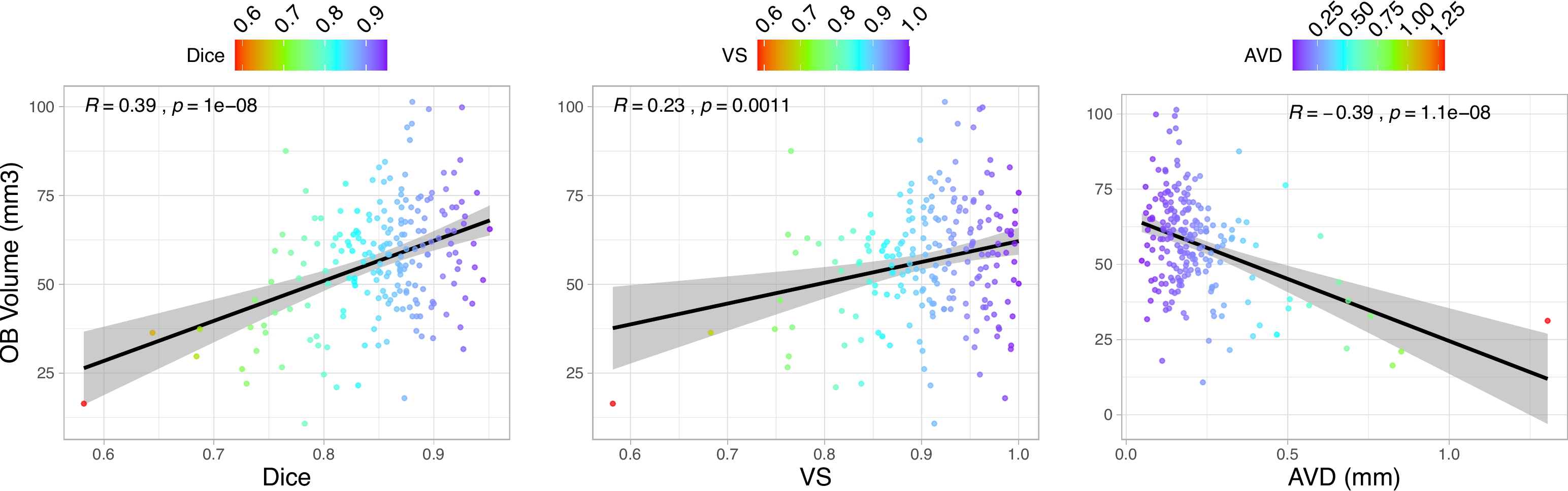}
    \caption{ Scatterplots of OB volume estimates and segmentation similarity metrics on the in-house test-set as well as the Pearson correlation coefficient and linear regression. We observed that segmentation performance decreased with OB size. Especially in subjects with a total OB volume smaller than 20~$mm^{3}$.}
    \label{fig:vol_vs_metric}
\end{figure*}

\begin{figure*}[!hbt]
    \centering
    \includegraphics[width=0.9\textwidth]{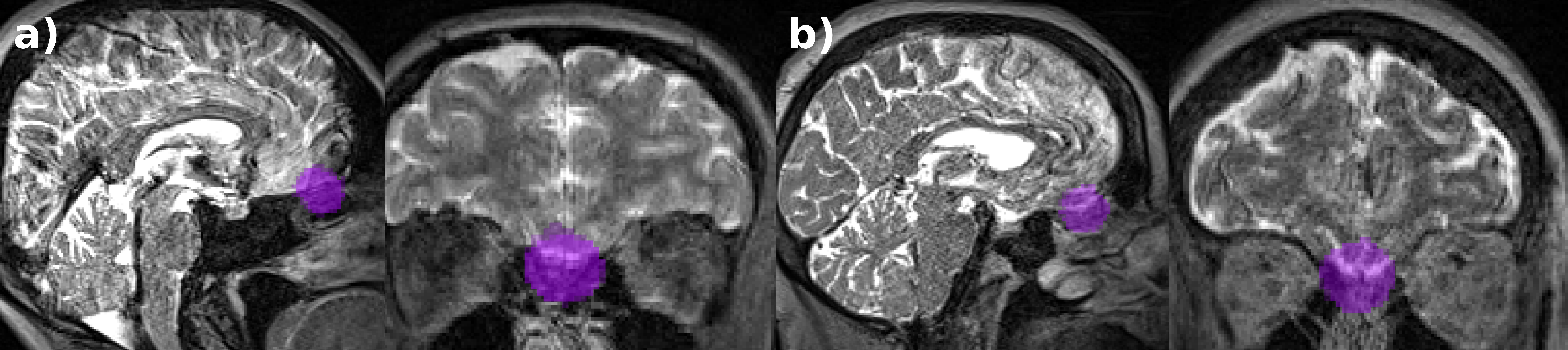}
    \caption{ Sagittal and Coronal T2-weighted MR images and predictions from the localization stage (purple) on two cases from the Rhineland Study. A-B) Present subjects excluded from the volume estimates sequence stability analysis (E5) due to severe motion artefact. Nonetheless, the localization stage still can detect a region containing both OBs.}
    \label{fig:motion}
\end{figure*}

\end{document}